\documentclass[journal,transmag]{IEEEtran}

\usepackage[section]{placeins}
\usepackage{subfigure}
\usepackage{caption}
\usepackage{graphicx}
\usepackage{lscape}
\usepackage{xcolor}

\usepackage{enumitem}

\usepackage[hidelinks]{hyperref}

\usepackage{lscape}
\usepackage{algorithm,algorithmic}
\usepackage{lineno}
\usepackage{amsmath}
\usepackage{empheq}

\usepackage[noadjust]{cite}

\newcommand{\xx}{{congested network property }}
\newcommand{\xxc}{{congested network property}}
\newcommand{\ta}{{\sc REINFORCE-TD}}
\newcommand*{\fracjl}[2]{#1/#2}
\newcommand{\blueOLD}[1]{\textcolor{black}{#1}}
\newcommand{\blue}[1]{{\color{black}#1}}
\newcommand{\minor}[1]{{\color{black}#1}}

\newcommand{\RoundTwo}[1]{\textcolor{black}{#1}}

\ifCLASSINFOpdf
\else
\fi

\begin{document}

\title{Congested Urban Networks Tend to be Insensitive to Signal Settings: Implications for Learning-Based Control}

\author{\IEEEauthorblockN{Jorge Laval\IEEEauthorrefmark{1}, and 
Hao Zhou\IEEEauthorrefmark{1}}
\IEEEauthorblockA{\IEEEauthorrefmark{1} School of Civil and Environmental Engineering,
Georgia Institute of Technology, Atlanta, GA 30332 USA}
\thanks{Corresponding author: J. Laval (email: jorge.laval@ce.gatech.edu).}}

\IEEEtitleabstractindextext{%
\begin{abstract}
This paper highlights several properties of large urban networks that can have an impact on   machine learning methods applied to traffic signal control. 
In particular, we note that the average network flow tends to be independent of the signal control policy as density increases past the critical density. We show that this property, which so far has remained under the radar, implies that 
no control (i.e. a random policy) can be an effective control strategy for a surprisingly large family of networks, especially for networks with short blocks.
We also show that this property makes deep reinforcement learning (DRL) methods ineffective when trained   under congested conditions\blue{, independently of the particular  algorithm used}.  
Accordingly, in contrast to the conventional wisdom around learning-based methods promoting the exploration of all states, we find that for urban networks it is  advisable to discard any congested data when training, and that doing so will improve  performance under all traffic conditions. Our results apply to all possible grid networks thanks to a parametrization introduced here. The impact of the turning probability was found to be very significant, in particular to explain  the loss of symmetry observed in the macroscopic fundamental diagram of the networks, which is not captured by existing theories that rely on corridor approximations  without turns. Our findings also suggest that  supervised learning methods have enormous potential as they require very little examples to produce excellent policies.
\end{abstract}

}

\maketitle

\IEEEdisplaynontitleabstractindextext

\IEEEpeerreviewmaketitle

\section{Introduction}

Congested urban networks are known to behave chaotically and to be very unpredictable, with \blue{microscopic} outputs   being hypersensitive to the input demand
\cite{daganzo1996nature,daganzo1998queue,nair2001non,adewumi2016application}. 
\blue{An explanation for these unpredictable dynamics has long been conjectured \cite{nagel1995emergent,nagatani2002physics,helbing2001traffic,chowdhury2000statistical,nagel2003still} based on the analogy of gas-liquid phase transitions \cite{stauffer2018introduction}. In this analogy, near the critical density chaotic dynamics emerge as a result of the power-law distribution of congested clusters--the area and the time-space plane where vehicles are stopped. Power laws are the hallmark of fractal objects and complex systems having phase transitions, and due to their infinite variance are responsible for the  chaotic and unpredictable dynamics \cite {schroeder2009fractals}. 
This complexity has proven hard to tackle in the literature, where  numerous signal control algorithms, mathematical programs and learning-based  control methods to optimize network performance have shown only mild success} \cite{khamis2014adaptive, chu2016large,xu2018network,ge2019cooperative,wei2019colight,tan2019cooperative,gong2019decentralized}. Although operational improvements have been shown in these references, they mostly correspond to light traffic conditions or very   small networks. And success has been limited when it comes to outperforming greedy benchmarks even under lightly congested conditions \cite{belletti2017expert}. 

But \blue{on large networks and on a coarser network-level scale} the  empirical verification of the existence of a network-level Macroscopic Fundamental Diagram (MFD) is a statement of the ``order emerging from chaos'' so ubiquitous in complex dynamical systems.
The MFD
gives the average link flow on a network as a function of the average link density, arguably independently of trip origins and destinations, and route choice. 
This robustness strongly suggests that the effects of signal timing  might be limited.
We argue here that the main cause of our struggles to control \blue{large} congested urban networks is not complexity but simplicity.  While small networks might exhibit high throughput variations \cite{daganzo1998queue}, we show here that large congested networks tend to produce a throughput that is quite predictable, and which cannot be controlled in congestion because it tends to be independent of the signal control policy. We call this property the \textbf{\textit{``\xxc''}}\blue{, and it is the main motivation of this paper to examine its consequences for traffic control}.

An early indication of this property, which unfortunately remained under the radar all these years, can be traced back to the work of Robert Herman and his group in the mid-eighties around the two-fluid model \cite{Herman84, Hani85, mahmassani1990network}. 
Although not mentioned explicitly in these references, some of the figures  reveal  a significant insensitivity of network performance with respect to signal control \footnote{For example, figures 9a, 10a and 11a in \cite{Hani1985} show that providing signal progression does not significantly improve the average speed on the network; figures 5.10 and 5.13 in \cite{Hani85} show that signal offset has no discernible influence on network throughput.}.  
They also find that  block length positively influences network throughput.
Notice that these results are based on the microscopic model NETSIM over an idealized grid network with identical block lengths, which might explain why the authors were reluctant to highlight the insensitivity to traffic signal control. 
Here we show that the \xx applies even for networks with  different block lengths and signal settings. 

A more recent indication of this property can be found in the stochastic method of cuts \cite{Laval2015Stochastic}, showing that the network MFD can be well approximated by a function of only three measurable network parameters: 
\begin{subequations}\label{parameters}
\begin{align}
	\lambda &\propto\fracjl{E(\mbox{block length})}{E(\mbox{green time})},\label{lambda0}\\
	\delta &=COV(\mbox{block length})\quad  \\ 
	\rho &= \fracjl{E(\mbox{red time})}{E(\mbox{green time})}
\end{align}
\end{subequations}
where $E(\cdot), COV(\cdot)$ stand for expectation and coefficient of variation, respectively, and block length refers to the distance between consecutive traffic lights.
But $\rho = 1$ when we consider all travel directions on grid networks, such as in this paper, and therefore can be dropped from the formulation. This strongly suggests that signal timing might be irrelevant when considering all directions of travel and only affects network performance by the average green time across all directions of travel.

Other recent indications of the \xx are hard to find, to the best of our knowledge.  With the exception of \cite{gayah2014impacts}--who found locally adaptive signals   to have little effect on the MFD in heavily congested networks--we conjecture that this is because the \xx is easy to miss unless (i) the space of all possible large-scale networks is explored (i.e., networks with different parameters $\lambda$ and turning proportion), and perhaps more crucially, unless (ii) the performance is analyzed under all density levels, e.g. with the MFD. As mentioned in the first paragraph, a single, small network is used in most studies, which prevents any general conclusions. In addition, typically performance is evaluated for prescribed demand patterns that do not span the whole range of densities. Even studies based on the MFD have not observed this property.   \cite{girault2016exploratory} studied the impacts of several control strategies on the MFD of an idealized grid network, and found that the impacts of coordination were highly sensitive to the signal cycle time, and that poor coordination can significantly decrease the network capacity and free-flow travel speed. Similarly, \cite{abdelghaffar2019novel} reported positive simulation results from signal control over a large network of Los Angeles (CA) downtown area  with 420 intersections. 
The most likely explanation for these observations is that  long-block networks were used; in fact, a quick sample of the block length for this latter study reveals that it is around 300 meters, which would imply a long-block network for an average green time (not reported in the study)  below 70 seconds, which appears very plausible.

The lack of awareness of the \xx is having negative consequences on emerging control technologies\blue{, and preventing this is the main motivation of the paper.} A case in point is deep reinforcement learning (DRL), where, we claim, potentially all training methods proposed  to date are unable to learn effective control policies as soon as congestion appears on the network. In fact, two decades ago \cite{camponogara2003distributed}  showed that training above the critical density the resulting DRL policies deteriorate significantly. While the current explanation of these observation is the potentially non-stationary and/or non-Markovian behavior of the environment \cite{choi2000hidden, da2006dealing}, here we posit that the congested network property\minor{,} and \textit{not} the particular DRL method used, is entirely to blame.


We also show that the turning probability at intersections is also a key variable that significantly affects the MFD.
Unfortunately, its impacts are not well understood in the literature as existing MFD estimation methods are based on arterial corridors without turning movements, and have been used as an approximation to the MFD of the whole network. This implies assuming that turning movements do not affect the MFD, which is not always the case. \cite{daganzo2011macroscopic} found the  time for a one-way street network seems to be insensitive to the probability of turning, but \cite{jin2013kinematic} found random turning ratios lead to more symmetric traffic patterns and higher flow-rates, and \cite{gayah2014impacts} found that a fixed turning probability of 0.2 leads a one-way street network towards gridlock more quickly than with random turning. A recent study  \cite{xu2020analytical} adopted a double-ring network approach to estimate analytically the impacts of turning on the MFD using the stochastic method of cuts. They found only slight variations in the MFD as a function of the turning probability.

In this paper we   
(i) provide additional evidence for the \xx by expanding the experiments in \cite{Herman84, Hani85, mahmassani1990network} to all grid network topologies, 
(ii) unveil additional properties such as loss of symmetry, overlapping and detaching,  and 
(iii) analyze how these properties affect the performance of machine learning methods applied to signal control using both the kinematic wave model (in the main text) and the off-the-shelf simulation model SUMO \cite{SUMO2012} (in the appendix). 
\minor{Towards} this end, the remainder of the paper is organized as follows. We start with the background section on the MFD, DRL and  a survey of related work. Then, we define the problem setup and apply it to a series of experiments that highlight the main properties found here. Finally, the paper concludes with a discussion and outlook section.

\section{BACKGROUND}

\subsection{\blue{The Macroscopic Fundamental Diagram (MFD)}}

The network MFD has its origins in \cite{godfrey1969mechanism} as a way of describing the traffic flow of urban networks at an aggregate level, and has been used in the past as a concise way of displaying network simulation output \cite{Smeed1967Road,Herman1979Two,Herman84, mahmassani1984investigation}. 
For a given traffic network, it  describes the relationship between traffic variables averaged across all \textit{lanes} in the network. 
The main requirement for a well-defined MFD is that congestion be homogeneously distributed across the network, i.e. there must be no ``hot spots'' in the network. For analytical derivations it is often also assumed that each \textit{lane} of the network obeys the  kinematic wave model \cite{Lighthill1955Kinematic,richards1956shock} with common fundamental diagram \cite{daganzo2008analytical, Laval2015Stochastic}. 
In this way, upper bounds for the MFD have been found using the method of cuts in the case of homogeneous networks \cite{daganzo2008analytical}. In a flow-density diagram a ``cut'' is a strthisaight line with slope (wave speed) corresponding to the average speed of a moving observer and intercept given by the maximum passing rate the observer would measure. By varying the observer speed one obtains a series of cuts whose lower envelope gives an approximation of the MFD. For general networks,  \cite{Laval2015Stochastic} introduces the stochastic method of cuts and shows that (the probability distribution of) the MFD can be well approximated by a function of the three  parameters in \eqref{parameters}. In particular to this paper,   the cuts' wave speed produced by this theory for extreme free-flow, $u_0$, and extreme congestion, $-w_0$, are given by:
    \begin{equation}\label{u0}
        u_0=w_0=\frac{4 \lambda }{\delta ^2+2 \lambda +1}.
    \end{equation}
after using $\rho= 1$  in equation (17b) of  \cite{Laval2015Stochastic}.

\blue{As we show in this paper, it turns out that the parameter $\delta$ that regulates the variance of block lengths affects the MFD only slightly, and thus can also be dropped from the formulation. The only variable left is $\lambda$, a measure of the propensity of the network to experience spillbacks, which  waste capacity. It was shown  in \cite{Laval2015Stochastic} that $\lambda< 1$ is the ``short-block condition'', i.e. the network becomes prone to spillback, which can have a severe effect on capacity. Conversely, a network with $\lambda>1$ has  long blocks (compared to the green time) and therefore will not exhibit spillback. \minor{We} show here that the \xx is more pervasive in networks with short blocks, where throughput tends to be insensitive to signal control for all density levels; in networks with long blocks this property tends to be observed in extreme free flow and extreme congestion only, leaving room for operational improvements around the critical density.
}

Notice that these estimation methods give the MFD of an arterial corridor without turning movements, and have been used as an approximation to the MFD of the whole network. This implies assuming that turning movements do not affect the MFD, which is not the case as will be shown here. \blue{Also note that the MFD will be used in this paper
simply to illustrate the results in the most concise manner possible, and does not intervene in generating these results.
}
 
\subsection{Reinforcement learning}

The use of deep neural networks  within Reinforcement Learning  algorithms has produced important breakthroughs in recent years. These deep reinforcement learning (DRL) methods have outperformed expert knowledge methods in areas such as arcade games, backgammon, the game  Go  and autonomous driving \cite{mnih2015human, silver2017mastering,chen2019model}. 
In the area of traffic signal control  numerous DRL control methods have been proposed both for isolated intersections \cite{li2016traffic,genders2016using} and small networks \cite{chu2015traffic,chu2019multi,tan2019cooperative,ge2019cooperative}.  The vast majority of these methods have been trained with a single (dynamic) traffic demand profile, and then validated using another one, possibly including a surge \cite{ge2019cooperative}. 

In the current signal control DRL literature the problem is treated, invariably, as an episode process, which is puzzling given that the problem is naturally a continuing (infinite horizon) one. Here, we adopt the \textit{continuing} approach to maximize the long-term average reward. We argue that in signal control there is no terminal state because the process actually goes on forever. And what may appear as a terminal state, such as an empty network, cannot be considered so because it is not achieved through the correct choice of actions but by the traffic demand, which is uncontrollable. An explanation for this puzzling choice in the literature might be that DRL training methods for episodic problems have a much longer history and our implemented in most machine learning development frameworks.
For continuing problems this is not unfortunately the case, and we propose here the training algorithm \ta, which is in the spirit of REINFORCE  with baseline \cite{willianms1988toward} but for continuing problems. To the best of our knowledge, this extension of REINFORCE is not available in the literature.

Reinforcement learning is typically formulated within the framework of a {\em Markov
decision process} (MDP). At discrete time step $t$ the environment is in state $S_t\in{\cal S}$, the agent will choose and action  $A_t\in{\cal A}$, to maximize a function of future rewards $R_{t+1}, R_{t+2}\ldots$ with $R_-: {\cal S} \times {\cal A} \rightarrow \Re$. There is a state transition probability distribution $\Pr(s',r|s,a)=\Pr(S_t=s',R_t=r|S_{t-1}=s, A_{t-1}=a)$ that gives the probability of making a transition from state $s$
to state $s'$ using action $a$ is denoted $\Pr(s,a,s')$, and is commonly referred to as the ``model''.
The model is
{\em Markovian} since  the state transitions are independent of any previous
environment states or agent actions. For more details on MDP  models the reader is referred to \cite{bellman1957markovian,bertsekas1987dynamic,howard1960dynamic,puterman1994markovian}

The agent's decisions are characterized by a stochastic
\textbf{policy} \( \pi (a|s) \), which is the probability of taking action
\( a \) in state \( s \).
In the continuing case the agent seeks to maximize the \textit{average reward}:
\begin{equation}\label{eta}
    \eta (\pi )\equiv \lim_{T\rightarrow\infty} \frac{1}{T}\sum_{t=1}^T E_{\pi}\left[R_t\right]
\end{equation}
 The term $E_{\pi}$ means that the expected value (with respect to the distribution of states) \RoundTwo{assuming} the policy is followed.

In the case of traffic signal control for large-scale grid network, methods based on transition probabilities are impractical because the state-action space tends to be too large as the number of agents increases.
An alternative approach that circumvents this  \textit{curse of dimensionality} problem---the approach we pursue here---are ``policy-gradient'' algorithms,  where the policy is parameterized as \( \pi (a|s;{\theta }), \theta \in \mathcal{R}^{m} \), typically a neural network. Parameters $\theta$ are adjusted to 
improve the performance of the policy $\pi$ by following the gradient of cumulative
future rewards, given by the identity
\begin{equation}
\label{policy_grad}
\nabla \eta=E_{\pi}[G_t \nabla_{\theta}\log  \pi (a|s)]
\end{equation}
as shown in \cite{sutton1999policy} for both continuing and episodic problems. In continuing problems cumulative rewards $G_t$ are measured relative to the average cumulative reward:
\begin{equation}\label{return}
    G_t=\sum_{i=t+1}^\infty (R_i-\eta(\pi))
\end{equation}
and is known as the \textit{differential return}.

\subsection{Related work}

\RoundTwo{Signal control has a long history in several research domains. It is beyond the scope of this study to survey all of the existing methods. Instead, in this section we will focus on the recent RL-based signal control methods that are applicable to congested large urban networks. }

The related literature is split between two approaches for formulating the large-scale traffic control problem, either a centralized DRL algorithm  or a decentralized method with communication and cooperation among multi-agents. 
The centralized approach \cite{genders2016using,li2016traffic,chu2016large}  adopts a single-agent and tries to tackle the high-dimensional continuous control problem by  memory replay, dual networks and advantage actor-critic \cite{lillicrap2015continuous, mnih2015human}. The decentralized method takes advantage of multiple agents and usually requires the design of efficient communication and coordination to address the limitation of partial observation of local agents. Current studies \cite{khamis2014adaptive,wei2019colight,tan2019cooperative,gong2019decentralized} often decompose the large network into small regions or individual intersections, and train the local-optimum policies separately given reward functions reflecting certain level of cooperation with neighboring agents. How to incorporate those communication information to help design the reward function for local agents remains an open question. 

The environment modeling, state representation and reward function design are key ingredients in DRL. For the environment emulator, most studies use popular microscopic traffic simulation packages such as  AIMSUM or SUMO. Recently, FLOW \cite{kheterpal2018flow} has been developed as a computational framework integrating SUMO with some advanced DRL libraries to implement DRL algorithm on ground traffic scenarios. \cite{vinitsky2018benchmarks} provided a benchmark for major traffic control problems including the multiple intersection signal timing. 
There also exist studies \cite{chu2015traffic,arel2010reinforcement,ge2019cooperative} adopting methods to use self-defined traffic models as the environment. Complementary to those microscopic simulation packages, macroscopic models are able to represent the traffic state using cell or link flows. The advantage of macroscopic models is twofold: i) reducing complexity in state space and computation ii) being compatible with domain knowledge from traffic flow theory such as MFD theory.

Expert knowledge has been included in some studies to reduce the scale of the network control problem. In \cite{xu2018network}, critical nodes dictating the traffic network were identified first before the DRL was implemented. The state space can be remarkably reduced. MFD theory cannot provide sufficient information to determine the traffic state of a network. For instance, \cite{chu2015traffic} successfully integrated the MFD with a microscopic simulator to constrain the searching space of the control policies in their signal design problem. They defined the reward as the trip completion rate of the network, and simultaneously enforcing the network to remain under or near the critical density. The numerical experiments demonstrated that their policy trained by the integration of MFD yields a more robust shape of the MFD, as well as a better performance of trip completion maximization, compared to that of a fixed and a greedy policy.

While most of the related studies on traffic control only focus on developing effective and robust deep learning algorithms, few of them have shown traffic considerations, such as the impact of traffic density. The learning performance of RL-based methods under different densities have not been sufficiently addressed. To the best of our knowledge, \cite{camponogara2003distributed} is the only study which trained a RL policy for specific and varied density levels, but unfortunately their study only accounted for free-flow and mid-level congestion. 
\cite{dai2011neural} classified the traffic demand into four vague levels and reported that inflow rates at 1000 and 1200 veh/h needed more time for the algorithm to show convergence. But they did not report network density, nor try more congested situations nor discussed why the converging process has been delayed. 
Most studies only trained RL methods in non-congestion conditions, \cite{ge2019cooperative} adopted the Q-value transfer algorithm (QTCDQN) for the cooperative signal control between a simple $2\minor{\times}2$ grid network and validated the adaptability of their algorithm to dynamic traffic environments with different densities, such as \RoundTwo{the} recurring congestion and occasional congestion. 

It can be seen that most recent studies focus on developing effective and robust multi-agent DRL algorithms to achieve coordination among intersections. The number of intersections in those studies are usually limited, thus their results might not apply to large networks. Although the signal control is indeed a continuing problem, it has been always modeled as an episodic process. From the perspective of traffic considerations, expert knowledge has only been incorporated in down-scaling the size of the control problem or designing novel reward functions for DRL algorithm. Few studies have tested their methods on a full spectrum of traffic demands, the learning performance under different traffic densities, especially the congestion regimes, has not been fully explored. 

\RoundTwo{It is worth noting that, based on the recent development of the MFD theory in the traffic flow domain, many perimeter and boundary signal control methods have been developed to reduce congestion. Relevant to our research, \cite{geroliminis2012optimal, haddad2012stability,aboudolas2013perimeter} proposed and investigated the perimeter control method, whose general idea is to control the inflow/outflow across city zones by adjusting the signals at the boundaries such that the congestion level in certain zones can be managed. 
The MFD control method is novel and promising. It well incorporates the expert knowledge from the traffic flow theory, and it is also centralized, which significantly differs from the decentralized signal control popular in the learning-based literature. To this end, we believe the MFD control methods suggest a promising research direction, which is to better leverage traffic flow models, and use cooperative mult-agent reinforcement learning (MARL) to better explore the learning potential. MARL is a heated research direction in computer science, and is certainly beyond the scope of this paper. Instead, we will limit the learning-based signal control methods to those who do not require a centralized controller or communications from neighbors, such that all signals in the network apply the same policy. }




\section{Problem definition}

\blue{This section formalizes all the ingredients needed to formulate our problem, including the traffic flow model to describe the behavior of vehicles, the type of network, the vehicle routing assumptions and the traffic signal configuration. 
}

\begin{figure}[tb]
\centering
\includegraphics[width=.8\linewidth]{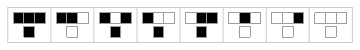}
\caption{CA Rule 184: The top row in each of the eight possible cases  shows the neighborhood values $( c_{i-1} ,  c_i , c_{i+1})$  and the bottom row \blue{shows} the updated $c_i$. Traffic flows from left to right in each of the 8 panels.}
\label{f1}
\end{figure}

\textbf{The traffic flow model} used in this paper is the kinematic wave model \blue{(or LWR model) } \cite{Lighthill1955Kinematic,richards1956shock}  with a triangular flow-density fundamental diagram \blueOLD{\cite {newell2002simplified}}. \blue{This  is the simplest model able to predict the main features of traffic flow, and it has become the standard analysis tool and traffic flow theory. 
It turns out that there are several models in the literature that are equivalent to the kinematic wave model: \minor{the} celebrated Newell's car-following model \cite{newell2002simplified} is the car-following version, which  can be formulated as a cellular automaton (CA) model \cite{Dag04a}.
But \cite{Lav16} showed that the shape of the triangular fundamental diagram  is irrelevant  thanks to a symmetry in the kinematic wave model,  allowing us to use an isosceles fundamental diagram, which has many useful properties in practice; see \cite{Lav16} for the details.  
This implies that
elementary CA rule 184 \cite{wolfram1984cellular} is in turn equivalent to the kinematic wave model, and will be used in this paper. Accordingly, here } we set  both the free-flow speed and the wave speed   equal to 1, implying  that the saturation flow is 1/2, the critical density $k_c$ is also 1/2 and the jam density 1, without loss of generality \footnote{ \blueOLD{The transformed densities used in this paper, $k$, are related to the ``real'' densities, $k'$, by $k'=
    2k/(\theta+1),  k<1/2$ and $k'=   1-2(1-k)\theta/(\theta+1),  k\ge 1/2$, where $\theta$ is there observed free-flow speed to wave speed ratio.
}}.
In a CA model,  each lane of the road is divided into  small cells  $i=1,2,\ldots \ell$ the size of a vehicle jam spacing, where cell $\ell$ is the most downstream  cell of the lane. The value in each cell, namely $c_i$, can be either ``1''  if a vehicle is present and ``0'' otherwise. The update  scheme  for CA Rule 184, shown in Fig.~\ref{f1}, operates over a neighborhood   of length 3, and can be written as:
\begin{equation}\label{CA Rule 184}
c_i :=c_{i-1}\lor c_{i-1}\land c_i \lor c_i\land c_{i+1} 
\end{equation}
The vector $c$ is a vector of bits  and \eqref{CA Rule 184} is Boolean algebra, which explains the high computational efficiency of this traffic model. \blue{ Notice that CA Rule 184 implies the exceedingly simple traffic rule ``advance if you can'', which can be understood as the canonical rule for traffic flow.} This also implies
that the current state of the system is described completely by the state in the previous time step; i.e. it is Markovian and deterministic. Stochastic components are added by the signal control policy, and therefore our traffic model satisfies the main assumption of the MDP framework. 

\begin{figure}[tb]
\centering
\includegraphics[width=.8\linewidth]{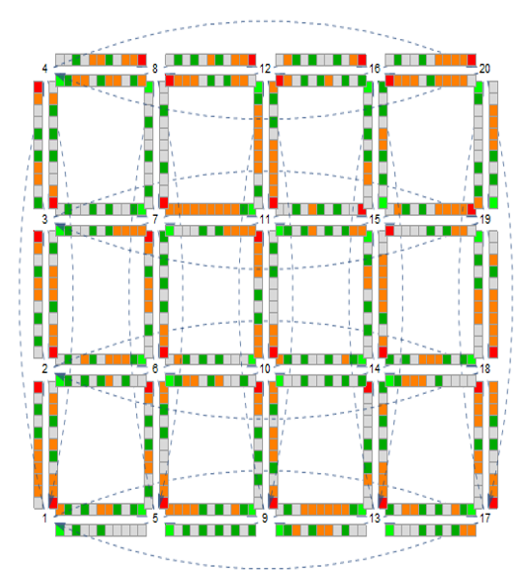}
\caption{Example  $4\times $5  traffic network. The connecting links to form the torus  are shown as dashed directed links; we have omitted the cells on these links to avoid clutter. Each segment has $\ell=10 $ cells;  an additional   cell has been added downstream of each segment to indicate the traffic light color.}
\label{network}
\end{figure}

\textbf{The network} corresponds to a grid network of  bidirectional streets with one lane per direction and with a traffic light on all intersections. To attain spatial homogeneity, the network is defined on a torus where each street can be thought of as a ring road where all intersections have 4 incoming and 4 outgoing approaches; see Fig.~\ref{network}.
\blueOLD{Let us define the ``N-S axis'' as the set of N-S bidirectional streets; similarly for the ``E-W axis''.}

\textbf{Vehicle routing} is random:
A driver reaching the stop line, say Mary,  will choose  to turn with probability $p$  or keep going straight with   probability $1-p$. If Mary decides to turn, she will turn left, right or U-turn with equal probability. For instance, $p=3/4$ gives an equal probability of 1/4 to all possibilities and therefore promotes a uniform distribution of density on the network\footnote{\blueOLD{We have verified that our  results are independent whether or not one includes U-turns.}}. 
If two or more vehicles are bound for the same approach during a time step, the tie is broken randomly. If the downstream approach is blocked  then Mary will not move during that time step, and will repeat the same selection process during the next time step.

Notice that this random routing \blueOLD{can be viewed} as a simplified form of driver adaptation, which avoids unrealistic bifurcations in the MFD \cite{daganzo2011macroscopic}. \footnote{Bifurcation takes place when drivers cannot clear the intersection because the downstream link in their route is jammed. If the driver does not adapt and change her route, the jam propagates even faster, eventually leading to a deadlock, with a portion of the links in the network being jammed, and the rest being empty. With driver adaptation, however, the jam propagation is slowed down by distributing congestion more uniformly across the entire network.}
It also makes our results applicable only to  grid networks where both supply and demand are spatially homogeneous, e.g. where origins and destinations are uniformly distributed across the network, such as in a busy downtown CBD. In the appendix we have tried other assignment methods that do create bifurcation, showing that the results of the paper remain unchanged. 

In the spirit of our discussion surrounding \eqref{parameters} we parametrize the space of all grid networks by 
$\lambda$ and $\delta$. As such,    
\textbf{the block length}, defined as the distance (in number of cells) between two neighboring traffic lights on a given street, is a random variable with coefficient of variation $\delta$, while keeping its
mean, $\ell$, constant. 

Notice that we have verified that values of $\ell\ge6$ do not change simulation results.

\textbf{Traffic signals} operate under the simplest possible setting with only red and green phases (no lost time, red-red, yellow nor turning phases). All the control policies consider here are \textit{incremental} in the sense that decisions are taken every $g$ time steps, which can be interpreted as a minimum green time:
After the completion of each green time of length $g$, the controller decides whether to prolong the current phase or to switch light colors. 

The following signal control policies will be used as baseline in this paper:
\begin{enumerate}[itemsep=-1mm]
    \item LQF: ``longest queue first" gives the green to the direction with the longest queue; it is a greedy methods for the ``best'' control, 
    \item  SQF: ``shortest queue first", a greedy methods for the ``worst'' control,
    \item RND: ``random'' control gives the green with equal probability to both directions, akin to no control. 
\end{enumerate}
The motivation to include SQF is that any possible control method will produce performance measures in between LQF and SQF.

To achieve the $\lambda$-parametrization we note from \eqref{lambda0} the expected green time is $E(\mbox{green time})=\ell(1/u+1/w)/\lambda$,
where $(1/u+1/w)$ is the proportionality constant in \eqref{lambda0}, $u$ and $-w$ are the fundamental diagram free-flow speed and wave speed, respectively, which are both equal to 1 in our scaling, so:
\begin{equation}\label{g0}
    E(\mbox{green time})=2\ell/\lambda 
\end{equation}
But under incremental control the expected green time is unknown a priory; it can only be estimated after the simulation run. In particular, setting a minimum  green time $g$ in the simulation will yield $E(\mbox{green time})\ge g$ due to the possibility of running two or more minimum green phases in a row. We have verified that under LQF $E(\mbox{green time})\approx g$, which is as expected because after a discharge it is very unlikely that the same approach would have the largest queue. Under the random policy the number of minimum green phases in a row is  described by a Bernoulli process of probability one half, and therefore $E(\mbox{green time})\approx 2g$. With this, we are able to compare LQF and RND control for a given $\lambda$ by  setting a minimum  green time $g$ in the simulation as:
\begin{subequations}\label{g}
\begin{empheq}[left={g=\empheqlbrace\,}]{align}
    2\ell/\lambda & \qquad \text{for LQF} \label{ga}\\
    \ell/\lambda & \qquad \text{for RND} \label{gb}
\end{empheq}
\end{subequations}
Unfortunately, under SQF a $\lambda$-parametrization is not possible because $\lambda$ becomes ill-defined. As we will detail shortly, at a given intersection $E(\mbox{green time}) \rightarrow \infty$ for one direction and $E(\mbox{green time}) \rightarrow 0$ for the other; i.e., after a few iterations the signal colors  becomes permanent at all intersections. 
Although this behavior is clearly unpractical, it turns out that SQF is \blue{the} key in  understanding the behavior of DRL methods in congestion.

\section{Baseline experiments} 

In this section we perform a series of experiments to highlight important properties of urban networks under the baseline control policies defined above. We are interested on the steady-state MFD  these policies produce  when deployed to all intersections in the network, and for different parameters $\lambda, \delta$ and $p$. 
The MFD for each policy is  obtained by  simulating this policy for  constant
network densities $k\in (0,1)$ and reporting the average flow in the network after 4 cycles\footnote{\blueOLD{Note that flows will not vary for different aggregation intervals because our network is designed to be both temporally and spatially homogeneous.}}.  This process is repeated  50 times for each density value to obtain an approximate 90\%-probability interval (between the 5th and the 95th percentile) of the flow for each density value.   Based on our results and to facilitates this discussion, we argue that networks have 4 distinctive traffic states: extreme free-flow ($k<0.2$), moderate free-flow ($0.2<k<0.5$), moderate congestion ($0.5<k<0.8$) and extreme congestion ($k>0.8$). 
\footnote{\blueOLD{In terms of the ``real'' densities, $k'$, and using footnote 3 with $\theta=4$,} we would have approximately: extreme free-flow ($k'<0.1$), moderate free-flow ($0.1<k'<0.25$), moderate congestion ($0.25<k'<0.7$) and extreme congestion ($k'>0.7$).}

\begin{figure}[tb]
\centering
\includegraphics[width=.99\linewidth]{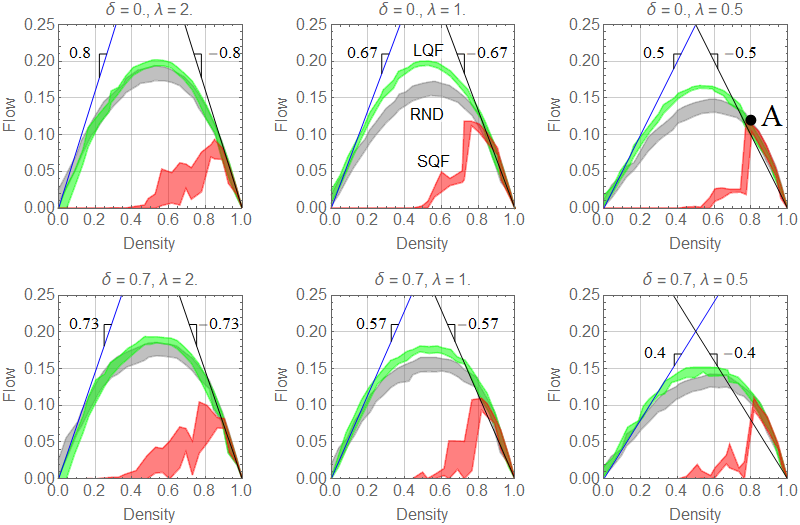}
\caption{Simulation results for baseline policies under equally-probable turning case $p=0.75$. 
First row: homogeneous networks (identical block lengths,  $\delta = 0$); second row: inhomogeneous  networks (highly variable block lengths, $\delta = 0.7$). The straight lines correspond to the extreme cuts 
in \cite{Laval2015Stochastic}, whose slopes $u_0$ and $-w_0$ are given by \eqref{u0}.
}
\label{MFDs}
\end{figure}

\begin{figure}[tb]
\centering
\includegraphics[width=.9\linewidth]{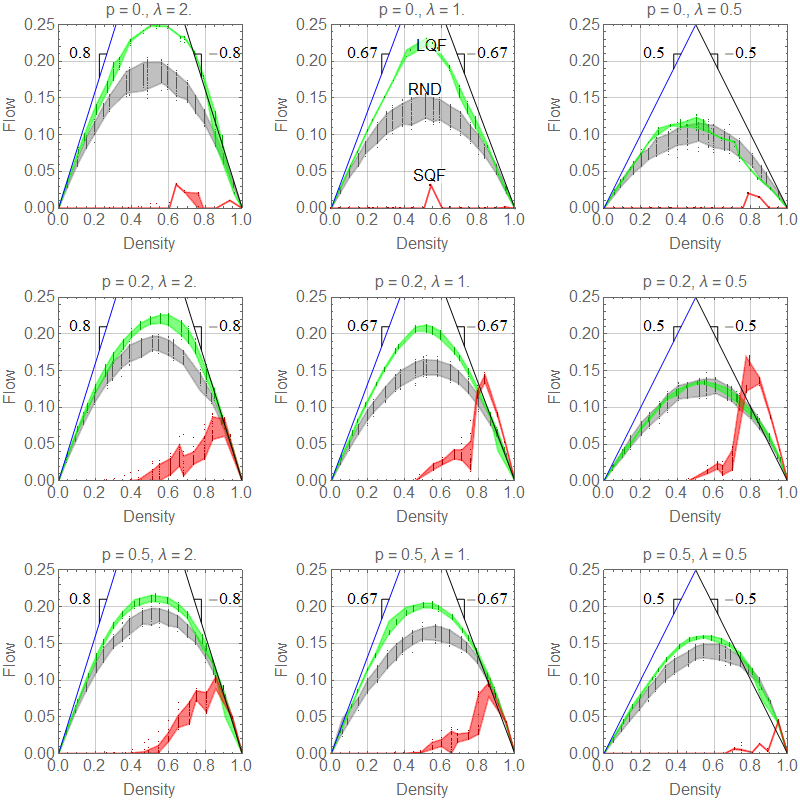}
\caption{Simulation results for baseline policies for homogeneous networks ($\delta = 0$) and different network parameter $\lambda$ (columns) and $p$ (rows). 
The straight lines correspond to the extreme cuts 
in \cite{Laval2015Stochastic}, whose slopes $u_0$ and $w_0$ are given by \eqref{u0}.
 Dots represent simulation points.
 }
\label{MFDsp}
\end{figure}

The following results are based on 2 figures showing our simulation results 
for  different network parameters $\lambda, p$ and $\delta$ for the three baseline signal control methods. Fig.~\ref{MFDs} shows the case of a fixed turning probability, $p=0.75$, and different network parameters $\lambda$ and $\delta$, to make the point that the effect of $\delta$ is small. 
The first row corresponds to homogeneous networks (identical block lengths,  $\delta = 0$), while the second row to inhomogeneous  networks (highly variable block lengths, $\delta = 0.7$).  
 Fig.~\ref{MFDsp} is for homogeneous networks ($\delta = 0$) and selected network parameter $\lambda$ and $p$, to see the impact of turning probabilities. These figures also show the extreme cuts produced by our earlier theory
in \cite{Laval2015Stochastic}, which correspond to straight lines of slopes $u_0$ and $w_0$ given by \eqref{u0}.

\begin{figure}[tb]
\centering
\includegraphics[width=.4\linewidth]{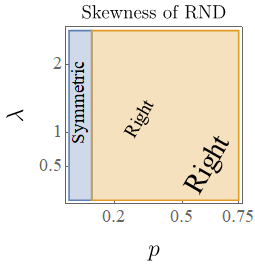}%
\qquad
\includegraphics[width=.4\linewidth]{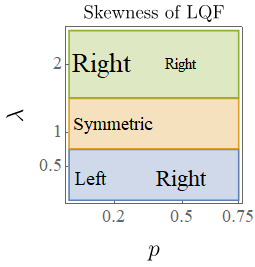}%

\includegraphics[width=.4\linewidth]{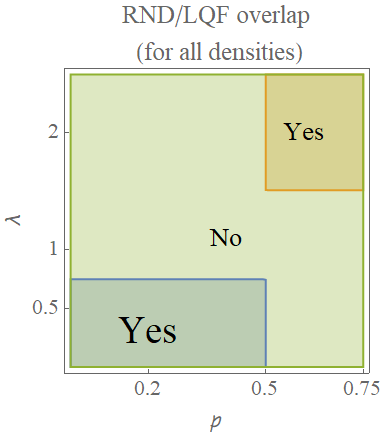}
\qquad
\includegraphics[width=.4\linewidth]{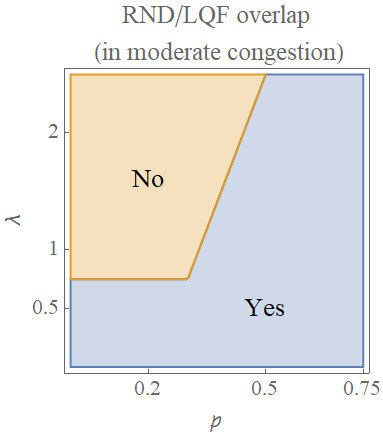}
\caption{Summary of results in the $(\lambda, p)$-plane. 
Top row:
Symmetric = no skewness,
Right =  skewed to the right,
Left =  skewed to the left.
Bottom left:
Yes = LQF  and RND overlap for all densities, 
No = LQF  and RND do not overlap for all densities.
Bottom right:
Yes = LQF  and RND overlap in moderate congestion, 
No = LQF  and RND do not overlap in moderate congestion. 
The font size is proportional to the effect being shown in each panel. These diagrams are approximate and were constructed by direct observation of a large number of flow-density charts such as the ones in Fig.~\ref{MFDsp}.
}
\label{BLsumm}
\end{figure}

\begin{figure}[tb]
\centering
\includegraphics[width=.45\linewidth]{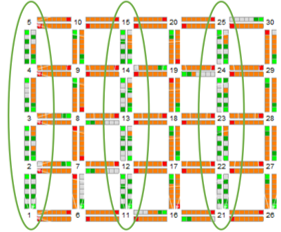}%
\qquad
\includegraphics[width=.4\linewidth]{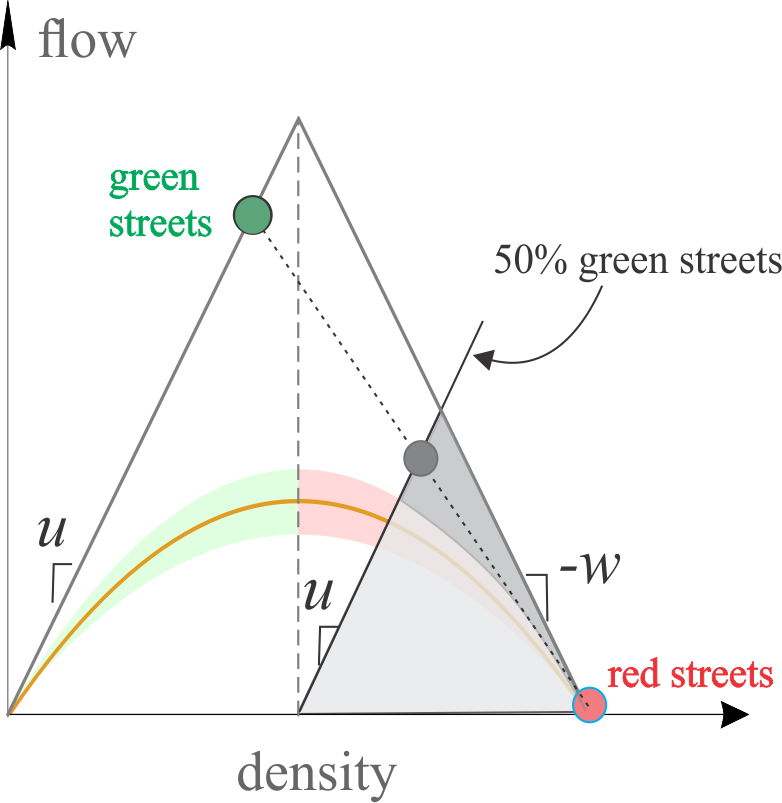}
\caption{SQF detaching. Left: the network reached an equilibrium where half of the N-S  streets are ``green steets'', and all other streets are ``red streets''. Right: low-density diagram showing  the high flows in the green streets (green disk), zero flow in all red streets (red disk) and the average traffic state in the network (gray disk). }
\label{f-detach}
\end{figure}

We can draw the following remarks from Fig.~\ref{MFDs} and \ref{MFDsp}:
\begin{enumerate}[label=\textbf{R-\arabic*},itemsep=2pt,parsep=2pt]
\item \label{lambda} \textbf{Effects of network parameters.} The block length parameter $\lambda$ has a significant impact on the MFD, especially for $\lambda<2$ and for the LQF and random control. This is as expected given that the probability of spillbacks in the network is directly related to $\lambda$. The variability of block length parameter  $\delta$  does not impact network throughput very significantly: by comparing the diagrams in each column of Fig.~\ref{MFDs} we can see that  the shape of the MFDs is practically the same, with inhomogeneous networks producing  slightly less capacity ($\approx 5\%$ on average). This result is surprising and indicates that networks with highly variable block lengths with mean $\ell$ performs  only slightly less efficiently than an idealised chessboard network with identical blocks of length $\ell$. To simplify our analysis in the sequel, we now assume  $\delta = 0$.

\item \label{loss of symmetry}  \textbf{Loss of symmetry.} There is a loss of symmetry in the MFDs for the LQF and RND policies, 
which can be traced to the turning probability $p$. From Fig.~\ref{MFDsp} we have mapped the level of skewness  in the $(\lambda, p)$-plane and summarize the results in the middle and right panels of Fig.~\ref{BLsumm}. It can be seen that the patterns are very different between the two policies, which is unexpected. Notice that for large turning probabilities the MFDs in congestion lie above the theoretical estimates, indicating that congestion propagates significantly faster due to turning movements.

\item \label{Detache} \textbf{Detaching: emergence of permanent street colors under SQF}. The middle row in Fig.~\ref{MFDsp} shows two examples where SQF flows exceed all other policies in congestion.
This ``detaching'' behavior in the MFD happens in $p\le 0.3$, and is a consequence of the signal color under SQF becoming permanent at each intersection. This induces a surprising collective pattern where all streets in the network are either under a permanent green  and at high flows, namely ``green streets'', or  under permanent red  and at zero flow, namely ``red streets''. All green streets belong to the same axis, say N-S,  which may contain some red streets; the other axis, E-W, contains only red streets.This is shown  in Fig.~\ref{f-detach} (left), where it can be seen that the network reached an equilibrium where half of the N-S  streets are green, and all other streets are red. Although permanent street colors tend to emerge for all values of $p$, we have observed that detaching only occurs in $p\le 0.3$, where the high flows in the green streets, shown as a green disk in the right side of the figure, are able to compensate for the zero flow in all red streets (red disk in the figure), such that the average traffic state in the network (gray disk) is above the LQF-MFD. 
Consideration shows that  depending on the proportion of green streets the 
average traffic state lays anywhere within the shaded triangle in the figure, whose left edge is achieved when
the proportion of green streets is maximal, i.e. 50 \%  in the case of square networks. Points along the line of slope $-w$ in the figure indicate that the N-S axis is operating in the congested branch of the FD, and points in the detaching area, that the N-S axis is operating in its free-flow branch. This is clearly unpractical but  indicates that a good strategy under severe congestion might be to favor one axis over the other.
\item \label{random}  \textbf{LQF/RND  overlap.} In a large number of networks the LQF  and RND   policies overlap over a significant range of densities, which
indicates that no control (i.e. RND) is an effective control method in such cases. This happens in extreme congestion and (to a lesser extent) in extreme free-flow on all networks. 
Consideration of (an extended version of) Fig.~\ref{MFDsp} reveals that the regions in the $(\lambda, p)$-plane    
where these policies overlap \textit{for all densities} can be summarized as in the bottom left panel of Fig.~\ref{BLsumm}. It can be seen that these regions represent a significant proportion of all possible grid networks.
\item \label{cnp} \textbf{The \xxc:} As density increases above  the critical density, network throughput  \textit{tends to be more and more independent} of  signal control, to the point of becoming absolutely independent of signal control once extreme congestion is reached. The precise  density at which this tendency is noticeable depends on the parameters  $\lambda$ and $p$, and can be approximated by the density at which LQF and RND first overlap in congestion. This is shown in the bottom right panel of Fig.~\ref{BLsumm}, which depicts the regions in the $(\lambda, p)$-plane where these policies first overlap in moderate congestion. It can be seen that the majority of networks are affected by this property.
\end{enumerate}

The precise mechanisms roughly outlined above are still under investigation and will be formulated on sequel papers. 
Here, we focus on the impact of the  \xx   in emerging control technologies, as it   shows that urban networks are more predictable than previously thought with respect to signal control. 
Recall from the introduction that earlier works from the late eighties found some evidence of the  \xx in \ref{cnp} using simulation on a homogeneous grid network;  studied the impacts of several control strategies on the MFD of an idealized grid network. We have also highlighted the importance of parametrizing urban networks by $\lambda$ and $p$ because of the unique and repeatable features that emerge under particular values of these parameters.

As we will see in the next section this \xx creates a challenge for learning effective control policies under congestion, where the policy tends to produce results similar to SQF, including detaching.

\section{Machine learning  experiments}

\begin{figure}[tb]
\centering
\includegraphics[width=.99\linewidth]{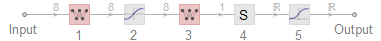}
\caption{Neural network architecture to approximate the policy. The numbers on top of the arrows indicate the dimensions of the corresponding input/output vectors, and the numbers below the squares are as follows: the input  is the state observable by the agent, 1: linear layer, 2: tanh function, 3: linear layer, 4: summation layer, 5: sigmoid function, and
 the output is a single real number that gives the probability of turning the light red for the N-S approaches.}
\label{f3}
\end{figure}

In this section we perform the same experiments in the previous section but with signal control policies based on machine learning methods. 
Each traffic signal is an agent equipped with a deep neural network with weights $\theta$ to represent the control policy  $\pi (a|s;{\theta })$, as shown in Fig.~\ref{f3}. It is a 3-layer perceptron with tanh nonlinearity,  known to approximate any continuous function with an arbitrary accuracy provided the network is ``deep enough'' \cite{kuurkova1992kolmogorov}. 
The input to the network is \textbf{the state observable by the agent} and corresponds to \textit{all} 8 approaches to-from the intersection: 
a vector of length 8, each entry representing the number of  vehicles in each approach. The output is a single real number that gives the probability of  turning the light red for the N-S approaches (and therefore turns the light green for the E-W approaches). Recall that these actions can be taken at most every $g$ time steps, per \eqref{g}.

Because our network is spatially homogeneous and without boundaries, there is no reason why policies should be different across agents, and therefore we will train \textit{a single agent} and share its parameters with all other agents. 
After training, we evaluate the performance of the policy  once deployed to all agents 
by observing the resulting MFD. 
Shown in all flow-density diagrams that follow is the mean LQF policy, depicted as a thick dashed curve. In this way, we are able to test the hypotheses that the policy outperforms LQF simply by observing if the shaded area is above the dashed line.
In particular, we will say that a policy is ``optimal'' if it outperforms  LQF, ``near-optimal'' if it performs similarly to LQF, and ``suboptimal'' if it underperforms LQF. 
To train the policy we will use the following 2 methods, each described the following subsections:
\begin{enumerate}[itemsep=-1mm]
\item Supervised learning: given labeled data, the weights are set to minimize the prediction error, and
\item Deep Reinforcement Learning (policy gradient). 
\end{enumerate}

\subsection{Supervised learning policies} \label{supervised section}

\begin{figure}[tb]
\centering
\includegraphics[width=.7\linewidth]{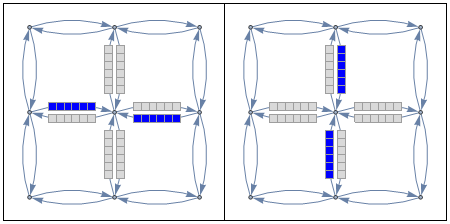}
\caption{Supervised learning experiment. Left: Extreme state $s_1$, where both N-S approaches are empty and the E-W ones are at jam density; we have omitted the cells on links other than the ones observable by the middle intersection to avoid clutter. Right: extreme state $s_2 $, the opposite of $s_1$.}
\label{supervised0}
\end{figure}

\begin{figure}[tb]
\centering
\includegraphics[width=\linewidth]{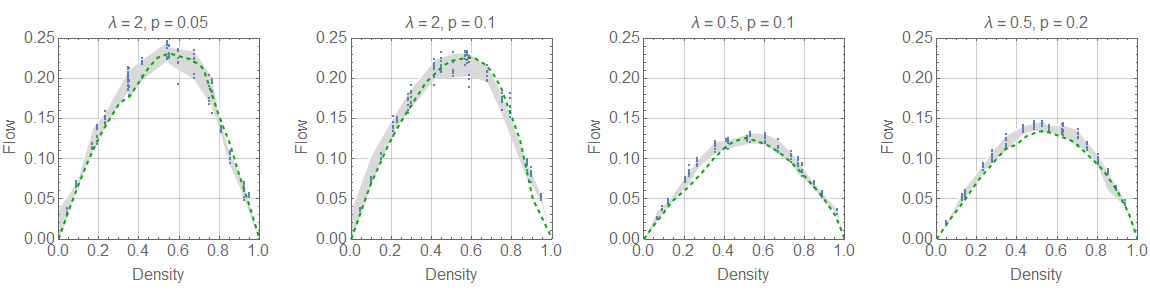}
\caption{Supervised learning experiment:  Resulting MFD (shaded area) for selected parameter values. Dots represent simulation points, and the dashed line the corresponding LQF-MFD.}
\label{supervised1}
\end{figure}

This section reports a rather surprising result: training the policy with \textbf{only two} examples yields a near-optimal policy. These examples are shown in Fig.~\ref{supervised0} and correspond to two extreme situations where the choice is trivial: the left panel shows extreme state $s_1$, where both N-S approaches are empty and the E-W ones are at jam density (and therefore red should be given to those approaches with probability one), while the right panel shows $s_2 $, the opposite situation (and therefore red should be given to N-S approaches with probability zero); in both cases all outgoing approaches are empty. The training data is simply:
\begin{equation}\label{trivial}
    \pi(s_1)\rightarrow 1,\qquad \pi(s_2)\rightarrow 0.
\end{equation}
We have verified that this policy gives optimal or near-optimal policies for all network parameters; see Fig.~\ref{supervised1}  for selected parameter values.

\subsection{DRL policies} 

Here the policy parameters are trained using DRL  on a single intersection using Algorithm \ref{reinforce} in the appendix. 
The density of vehicles in the network, $k$,  is kept constant during the entire training process. 
We define \textbf{the reward} at time $t,\ R_t,$ as the \textit{average advantage flow per lane}, defined here as the average flow through the intersection during $(t,t+g)$ \textit{minus} the flow predicted by the LQF-MFD at the prevailing density. 
In this context the LQF-MFD can be seen as a baseline for the learning algorithm, which reduces parameter variance. 

\begin{figure}[tb]
\centering
\includegraphics[width=\linewidth]{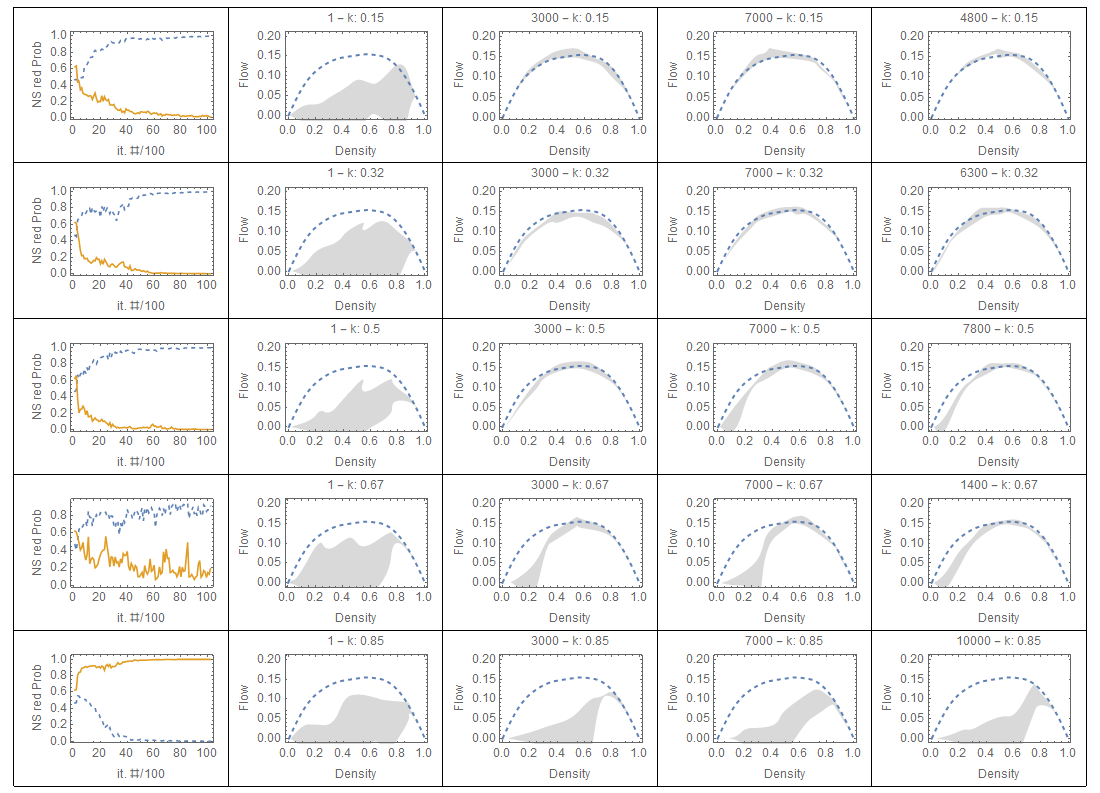}
\caption{Policies trained with constant demand and random initial   parameters $\theta$ with 
$\lambda=1/2$  and  equally likely turning probability $p=3/4$. The label in each diagram gives the iteration number and the constant density value.  First column: NS red probabilities of the extreme states, $\pi(s_1)$ in dashed line and $\pi(s_2)$ in solid line. The remaining columns show the flow-density diagrams obtained  at different iterations, and the last column shows the  iteration producing the highest flow at $k= 0.5$, if not reported on a earlier column.
\RoundTwo{Each row corresponds to a different density for training.}
}
\label{constantr}
\end{figure}

The results for network parameters $\lambda=1/2, p=3/4$, and
random  initial policy weights $\theta$ are shown in Fig.~\ref{constantr}. 
Each row corresponds to a constant training density $k$, while the first column depicts the NS red probabilities of the extreme states, $\pi(s_1)$ and $\pi(s_2)$ (described in \blue{Section} \ref{supervised section}) as a function of the iteration number, and these probabilities should tend to \eqref{trivial} for ``sensible'' policies. To facilitate the  discussion, let DRL-F be a DRL policy trained under free-flow conditions, i.e. $k< 0.5$ and DRL-C under congestion, i.e. $k\ge 0.5$. 
We can see from Fig.~\ref{constantr} that:
\begin{enumerate}[itemsep=-1mm]
    \item DRL-F policies are only near-optimal, with lower training densities leading to policies closer to LQF.

    \item  DRL-C  policies are suboptimal and deteriorate as $k$ increases. Sensible policies cannot be achieved for training density $k\ge 0.7$ as probabilities $\pi(s_1)$ and $\pi(s_2)$ converge to the SQF values; see the first column in Fig.~\ref{constantr}. 
    
\end{enumerate}

\begin{figure}[tb]
\centering
\includegraphics[width=.4\linewidth]{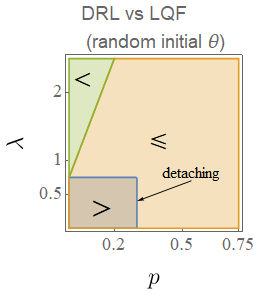}%
\qquad\qquad
\includegraphics[width=.4\linewidth]{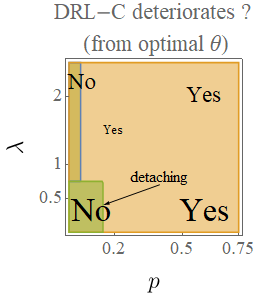}
\caption{Summary of results in the $(\lambda, p)$-plane. Left: regions where, starting from random initial parameters, the ``best'' DRL policy (across training densities)  performs better $(>)$, worse $(<)$ or slightly worse than $(\le)$ LQF. Right: regions where the DRL policy trained in congestion deteriorates, starting with optimal parameters. The font size is proportional to the effect being shown in each panel.}
\label{DRLsumm}
\end{figure}

\begin{figure}[tb]
\centering
\includegraphics[width=\linewidth]{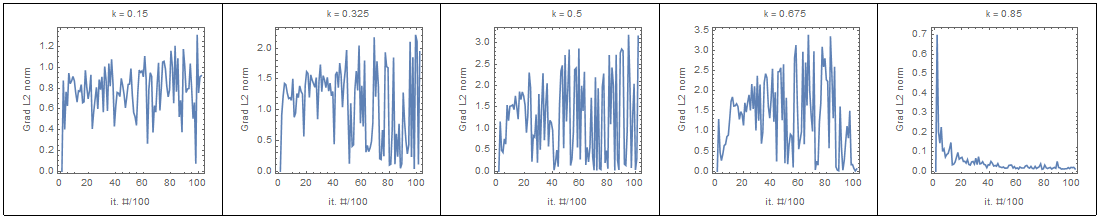}
\caption{Evolution of the gradient $L_2$-norm $||\nabla_{\theta}\log  \pi||_2$  as a function of the iteration number. The gradient goes to zero not because an optimal policy has been reached but because all actions yielded similar throughput near point ``A'' in Fig.~\ref{MFDs}.}
\label{grad}
\end{figure}

These observations indicate that DRL policies are only near-optimal and lose their ability to learn  a sensible policy as the training density $k$ increases. 
This is consistent with the \xxc, whereby the more the congestion, the less the policy affects   intersection throughput. 
This can make the DRL-C gradient $\nabla_{\theta}\log  \pi\rightarrow 0$ for  $p\ge 0.5$, not because an optimal policy has been reached but because there is nothing to be learned at that density level; see  Fig.~\ref{grad}.
Notice that  for lower turning probabilities the tendency of the  DRL-C gradient is similar to the other panels in the figure, but still the probabilities $\pi(s_1)$ and $\pi(s_2)$ converge to the SQF values.

While the above findings are true for all network parameters,  it is important to note that DRL can outperform LQF in some cases. In $0.05\le p\le 0.2, \lambda <1$ detaching is learned by (i) DRL-F with random initial weights, and by (ii) DRL-F and DRL-C with optimal weights given by supervised learning. That DRL-F can learn detaching is surprising since at that training density level ($k<0.5$) detaching does not take place. Not so for DRL-C because we have shown that it tends to SQF. 
The DRL detaching behavior is even more pronounced than SQF, which indicates that the number of ``green streets'' obtained by this policy is higher than under SQF. This improvement upon SQF is unexpected and indicates that DRL is able to learn how to outperform LQF in congestion, albeit impractically since signal colors become permanent.

\begin{figure}[tb]
\centering
\includegraphics[width= \linewidth]{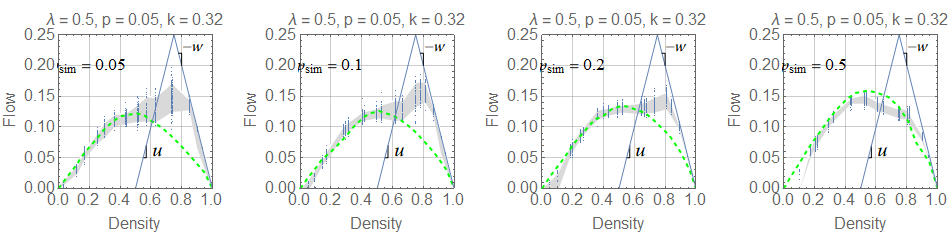}
\caption{Detaching policy found by DRL. Each panel shows simulation results with the same policy but with different turning probabilities during simulation, $p_{\text{sim}}$.
Dashed line: LQF-MFD for $\lambda=0.5, \delta=0$ and $p_{\text{sim}}$. The straight lines are included to facilitate the comparison with  our explanation in Fig.~\ref{f-detach}.}
\label{DRLdetach}
\end{figure}

This is shown in Fig.~\ref{DRLdetach}, where the first panel shows how the resulting MFD detaches from the LQF-MFD,  in a way consistent with our explanation in Fig.~\ref{f-detach}. 
The remaining panels show simulation results with the same policy but with higher turning probabilities during simulation, $p_{\text{sim}}$ in the figure. It can be seen that the detaching decreases with $p_{\text{sim}}$  eventually disappearing for $p_{\text{sim}}\ge 0.3$, at which point the network is subject to the \xxc.

Finally, Fig.~\ref{DRLsumm} shows a summary of results in the $(\lambda, p)$-plane. The left panel shows the regions where the DRL policy trained at any  density and starting from random initial parameters performs better $(>)$, worse $(<)$ or comparable to $(\approx)$ LQF. It can be seen that except for detaching, the DRL policy always underperforms LQF. 
The right panel shows the regions where the DRL policy trained in congestion deteriorates, starting with optimal parameters from the supervised experiment. \blue{In summary, except for detaching and when $p=0$, the additional DRL training under congested conditions leads to a deterioration of the policy, which increases with $p$. }

\section{Discussion and outlook} 

This paper has raised more questions than answers by exposing several important properties of urban networks that have remained unnoticed for decades, and that have important implications for traffic control. While a sequel paper will explore the theoretical aspects of these properties, here we focused on their impact on machine learning methods applied to traffic signal control on large networks. Although our results apply only to inhomogenous grid networks with uniformly distributed origins and destinations, we strongly suspect that the mechanisms unveiled here remain important driving forces in more general networks. For example, a key lesson is that urban networks need to be parameterized at least by $\lambda$ and $p$ before anything meaningful can be said about their performance. This consistent gap in the literature, which treats long- and short-block networks alike, may be responsible for the admittedly incremental advances of traffic  signal control in the last decades.

Our main result is the \xxc: on congested urban networks the intersection throughput tends to be independent of signal control. This property affects all types of signal control once the density exceeds the critical density, by rendering them more and more similar to random (i.e. no) control as density increases. 
The bottom right panel of Fig. \ref{BLsumm} \blue{shows} that  the MFD for  LQF  and RND policies start to overlap in moderate congestion in roughly 2/3 of the cases, which is an indication that the \xx applies to most urban networks. It is worth recalling that in severe congestion all policies (LQF, RND and SQF) overlap  on all networks, indicating that signal control has absolutely no effect on network throughput at those densities. Of course, this may be as expected since in extreme congestion all approaches are full; the LQF/RND overlap in other traffic states is less intuitive,  and remains an open question.

\subsection{Implications for DRL}

\subsubsection{Conjecture on the challenges faced by DRL}

We have seen that the \xx  hinders the training process under congested conditions
, which in some cases leads to learning the worst policy, SQF.
Even starting with initial weights given by the supervised training policy, we saw that additional training under congested conditions leads to a deterioration of the policy. We have verified similar behavior  under dynamic demand loads whenever congestion appears in the network.
This means, potentially, that all the DRL methods proposed in the literature to date are unable to learn sensible policies and deteriorate as soon as congestion appears on the network. It might also explain DRL's limited success  for traffic signal control  problems observed so far, currently believed to be due to urban networks being non-stationary and/or non-Markovian  \cite{choi2000hidden,da2006dealing}. We believe instead that the \xx is to blame, \blue{independently of the DRL method used (see appendix),} and that future work should focus on  new DRL methods able to extract relevant knowledge from congested conditions. 
In the meantime, it is advisable to train DRL policies under free-flow conditions only, discarding any information from  congested ones, as we have shown here that such free-flow DRL policies are comparable to LQF. 

A full explanation of the effects of the \xx on DRL is still missing. 
An important clue was provided earlier that the DRL-C gradient tends to vanish, not because an optimal policy was found but because there is nothing to be learned at that density level; see  Fig.~\ref{grad}. This is consistent with the chaotic nature of network traffic near the critical density mentioned in the introduction, and research is needed to test whether or not this is sufficient to explain the gradient behavior observed here.

\begin{figure}[tb]
\centering
\includegraphics[width= 0.9 \linewidth]{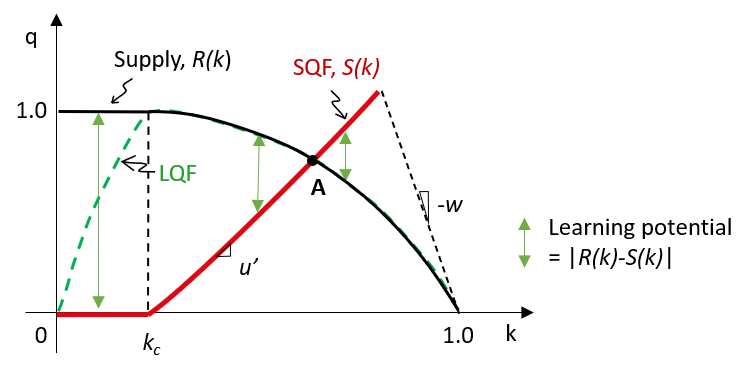}
\caption{DRL learning potential in real coordinates. Notice that the positive flows under the SQF policy shown correspond to a linear approximation with $u'<u$.}
\label{simple-mfd}
\end{figure}

We conjecture that SQF may provide additional insight, since its behavior  is markedly different in both traffic regimes; recall Figs.~\ref{MFDs} and \ref{MFDsp}.  An explanation consistent with our observations is presented in Fig. \ref {simple-mfd}, where it is conjectured that the ``learning potential'' at a given density is proportional to the gap, in absolute value,  between the network supply function under LQF and the flow under SQF: maximal in free flow, starts shrinking at the critical density $k_c$ to become negligible when reaching extreme congestion and finally increasing again as density keeps increasing.     
The precise shape of this diagram,  and the underlying impacts for traffic control, depends only on parameters $\lambda$ and  $p$, as expected. For example, on a long-blocks network the potential to the right of point ``A'' in the figure would be negligible, explaining why DRL does not learn detaching on these networks. But the shape of this diagram also depends on the flow under the SQF policy, which we strongly suspect is heavy-tailed. \blue{Additional research is needed to validate our conjecture.}

That DRL successfully learns LQF in free-flow conditions
indicates that most of what the agent needs to learn, e.g. flushing the longest queue, is encoded in free flow.
This might be intuitive considering that in free flow the reward tends to grow linearly with the state variables since there are no spillbacks. 
Less intuitive is that  DRL consistently learns SQF in congestion.  Since the gradient information tends to become less useful due to the \xxc, one would expect that it learns RND instead. Why this happens remains an open question, and might hold the key to improving DRL methods in moderate congestion. 

\subsubsection{Turning probabilities}
The impact of the turning probability $p$ turned out to be very significant, not only for explaining the behavior of baseline policies but also for endowing DRL to find policies that exceed SQF's ability to produce the detaching phenomenon introduced in this paper.
The case $p\rightarrow 0$ can be problematic in the current framework  because the environment tends to be deterministic, which  contradicts the assumptions of the type of stochastic gradient descent methods traditionally used in DRL. We observed that for $p<5\%$ near-optimal DRL policies are hard to find. 
Turning probabilities also explain the loss of symmetry observed for LQF and RND baseline policies, which is not captured by existing theories that rely on corridor approximations  without turns. Unveiling the mechanisms for the loss of symmetry due to turning should provide significant insight into the operation of urban networks. It becomes clear that future research should focus on mapping the origin and destination table and dynamic traffic assignment models to turning probabilities. \blueOLD{Future work should also explore whether a simple double-ring network is able to capture the main results in this paper, possibly extending the framework proposed in \cite {xu2020analytical}.}

\subsubsection{State representations}
Although not shown in the main text, we have verified that other state representations have little impact on the resulting machine learning policies. Besides from a vector of length 8 used in the main text as input to the neural net, where each entry represents the number of queued vehicles in each approach to/from the intersection, we also tried (i) a vector of length 4 only considering incoming approaches, (ii) a $8\times \ell$ matrix of bits, given the four incoming and the four outgoing $c$-vectors from the CA model, one for each approach to/from the intersection, and (iii) a $4\times \ell$ matrix only considering incoming approaches.
Considering all 8 approaches to/from the intersection would make it possible for the model to learn to avoid spillbacks. But according to the main result in this paper, this does not happen. Instead, we found that with the $8\times \ell$ input (but not the $4\times \ell$)  supervised learning yields a near-optimal policy with only two examples. This is surprising because the outgoing approaches are simply null vectors in these two examples used for training, but somehow the larger configuration endows the model with better extrapolation capabilities.

\subsection{The promise of supervised learning}
Notably, we also found that supervised learning with only two examples yields optimal or near-optimal policies for all network parameter values. This intriguing result indicates that  extreme states $s_1$ and $s_2$ encode vital information and that the neural network can successfully extrapolate to all other states. 
Understanding precisely why this happens could lead to very effective supervised learning methods based on expert knowledge, and perhaps to supplement DRL's inability to learn under congested conditions.

\subsection{Generality of our results}
A crucial assumption in this work was full driver adaptation to avoid bifurcations in the MFD, which have not been observed in the field to the best of our knowledge. In the appendix, we have verified the congested network property and its detrimental impact on DRL still hold true with more realistic routing behaviors and network configurations in SUMO, where vehicles have actual destinations and different levels of driver adaptation. The code is also shared in an open Github repository:\url{https://github.com/HaoZhouGT/signal\_control\_paper}

With non-adaptive drivers a different DRL framework would have to be used with the reward function having to capture the possibility of localized gridlocks and the state observable by the agent having to capture their spatial extent. The agent might also need to learn the origin-destination matrix since gridlock probabilities grow with the number of nearby origins and destinations. 
One of our ongoing studies \cite{zhou2021gridlock} is focusing on more realistic driver adaptation mechanisms. We conjecture that their impacts will be observed mainly on networks with short blocks under extreme congestion, where the blockage probability is higher, and that this will lead to bifurcations depending on the parameters of the driver adaptation model. 
 

\subsection{Implementation challenges of detaching}
Finally, detaching is a surprising finding that deserves more attention. Fig 6 provides a complete picture of the theory behind it, but its implementation might be controversial. One alternative might be favoring one axis over the other most of the time, but still giving the right of way to the other axis from time to time. This implementation challenge is the focus of future research by the authors. 

\section*{Acknowledgements}
The authors are grateful to Hani Mahmassani for pointing out the studies from the mid-eighties mentioned in the introduction. This study has received funding from NSF research projects \# 1562536 and \# 1932451 and by the TOMNET University Transportation Center at Georgia Tech.

\appendices
\section{The training algorithm \ta}
In this paper we propose the training algorithm \ta, which is in the spirit of REINFORCE with baseline \cite{willianms1988toward} but for continuing problems. To the best of our knowledge, this extension of REINFORCE is not available in the literature, which is almost entirely focused on episodic problems as discussed earlier. Notice that we tried other methods in the literature 
with very similar results, so \ta\  is chosen here since it has the fewest hyperparameters: learning rates $\alpha $ and $\beta$ for  weights,  $\theta $, and average reward, $\eta(\pi) $, respectively. Using a grid search over these hyperparameters resulted in $\alpha=0.2 $ and $\beta=0.05$.

Recall that  REINFORCE is probably the simplest policy gradient algorithm that uses \eqref{policy_grad} to guide the weight search. In the episode setting it is considered a Monte-Carlo method since it requires full episode replay, and it has been considered to be incompatible with continuing problems in the literature \cite{sutton2018reinforcement}. 
Here, we argue that a one-step Temporal Difference (TD) approach \cite{sutton1988learning} can be used instead of the Monte-Carlo replay to fit the continuing setting.
This boils down to estimating the differential return \eqref{return} by the temporal one-step differential return  of an action:
\begin{equation}\label{return2}
    G_t\approx R_t-\eta(\pi)
\end{equation}
Notice that the second term in this expression can be interpreted as a baseline in REINFORCE, which are known to reduce weight variance.
The pseudocode is shown in Algorithm \ref{reinforce}.

\begin{algorithm}
\caption{\ta}
\label{reinforce}
\begin{algorithmic}[1]
\STATE Input: weightized policy \( \pi (a|s;{\theta }), \theta \in \mathcal{R}^{m} \),  average density $k$
\STATE Set hyper-parameter $\alpha,\beta$, set average reward $\eta=0$
\STATE Initialize vector $\theta$ 
\STATE Initialize the network state $S$ as a Bernoulli process with probability $k$ over the cells in the network 
\REPEAT 
\STATE Generate action $A\sim \pi(\cdot|S;\theta)$ 
\STATE Take action $A$, observe the new state $S'$ and reward $R$ (by running the traffic simulation model for $g$ time steps)
\STATE $G\gets R-\eta$
\STATE $\eta\gets\eta+\beta \ G$
\STATE $\theta\gets\theta+\alpha\  G \ \nabla_{\theta}\log  \pi (A|S;\theta)$
\STATE $S\gets S'$
\UNTIL{forever}
\end{algorithmic}
\end{algorithm}

\section{Repeating DRL experiments in a microscopic simulator: SUMO}

We include this appendix to highlight that the results presented here can be replicated and extended using the  the open-source microscopic traffic simulator SUMO \cite{SUMO2012}. 

All DRL settings are kept as in the main text, except for the state representation, which corresponds here to a vector containing the queue lengths of each lane approaching the intersection. 
For training, we implemented both REINFORCE-TD and \blue{more advanced DRL methods for the continuing setting in \cite{sutton2018reinforcement}, including the well-known Actor to Critic (A2C). We also verified the learning challenge found by this paper is independent of the algorithm. As evidence we repeated the same DRL experiment using a more recent alternative to A2C, the Proximal Policy Optimization (PPO) \cite{schulman2017proximal} which is known to be more robust to gradient variances by constraining the update step using a similar idea of trust region optimization. The results of PPO method is shown in Fig.\ref{ppo-method}, where we tested the trained policy using PPO against a straightforward scenario where the queue only takes place at the east-west left-turn lanes and accordingly a sensible policy should choose to give green lights to east-west with a high probability close to 1. Fig.\ref{ppo-method} shows that the PPO algorithm successfully outputs a sensible policy when trained in the free flow state ($k=0.1$), but the learning fails to converge at high densities, e.g. ($k=0.7$). The results suggests that the learning challenge found in this paper should be independent of DRL algorithm.} The code is also shared in an open Github repository:\url{https://github.com/HaoZhouGT/signal\_control\_paper}

Fig \ref{sumo_mfd} shows the (untransformed) SUMO experiment results for the 3 baseline policies studied here and for a DRL policy trained under congested conditions (DRL-C) using REINFORCE-TD. It can be seen that these results are consistent with the model in the main text.

\begin{figure}[!htbp]
\centering
\includegraphics[width=\linewidth]{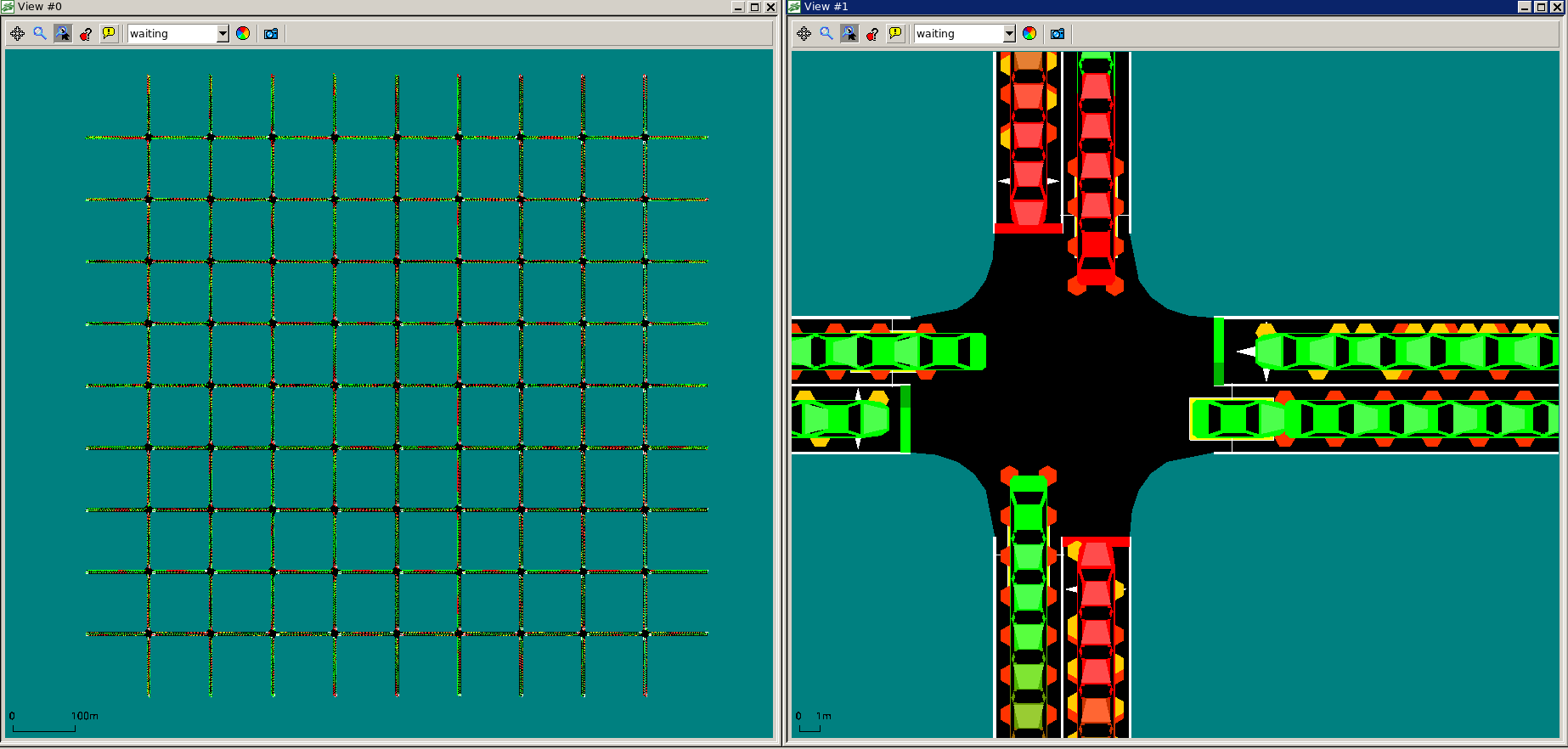}
\caption{A $9\times9$ grid network in SUMO: All roads are one-lane, traffic lights are two-phase without protected left-turns. $\lambda=4.0$ and $\delta= 0$. Vehicles randomly turn at intersections and make U-turn at boundaries. Notice that we modified the default routing algorithm in SUMO to achieve high density level without gridlock. Internal links of intersections are removed and downstream spill-back are prevented.}
\label{sumo_net}
\end{figure}

\begin{figure}[!htbp]
\centering
\includegraphics[width=1.0 \linewidth]{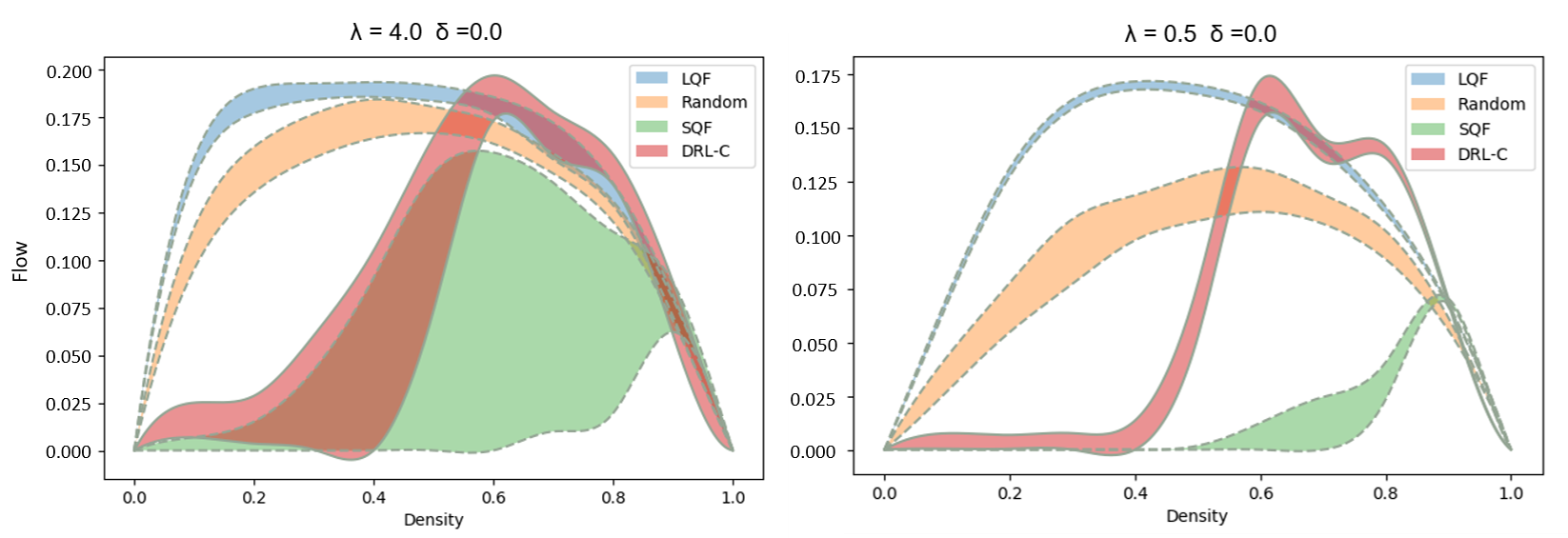}
\caption{MFD of DRL-C and baseline policies: Upper and lower envelopes correspond to the 95\% and 5\% percentiles of average network flow from 100 trails. Red shaded area depicts the MFD of DRL-C in extreme congestion (density 0.9). Curves are derived through interpolation from discrete point values at density 0.1,0.2,...,0.9. }
\label{sumo_mfd}
\end{figure}

\begin{figure}[!htbp]
\centering 
\subfigure[$k=0.1$]{
\includegraphics[width=0.45\linewidth]{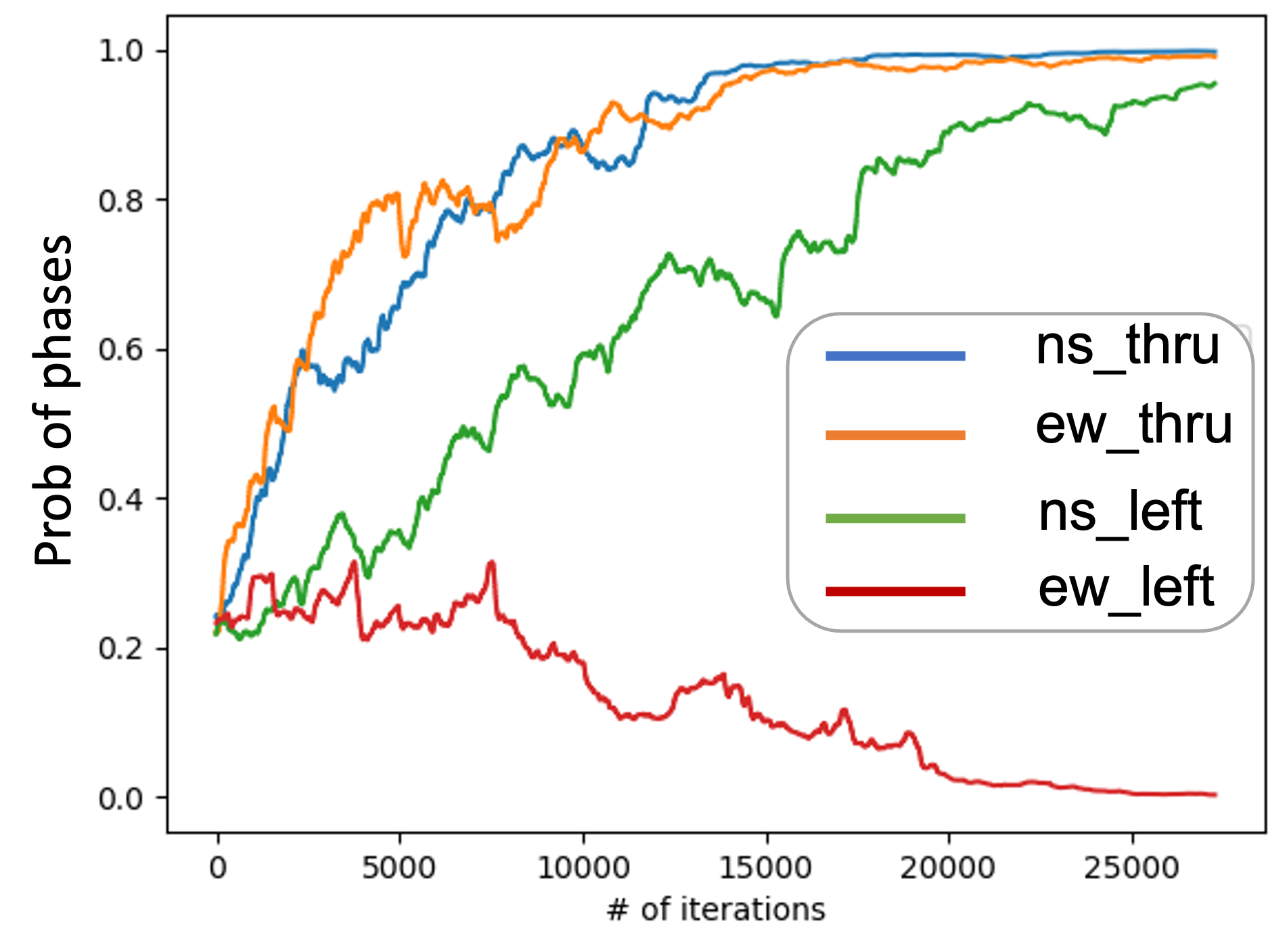}
}
\subfigure[$k=0.7$]{
\includegraphics[width=0.45\linewidth]{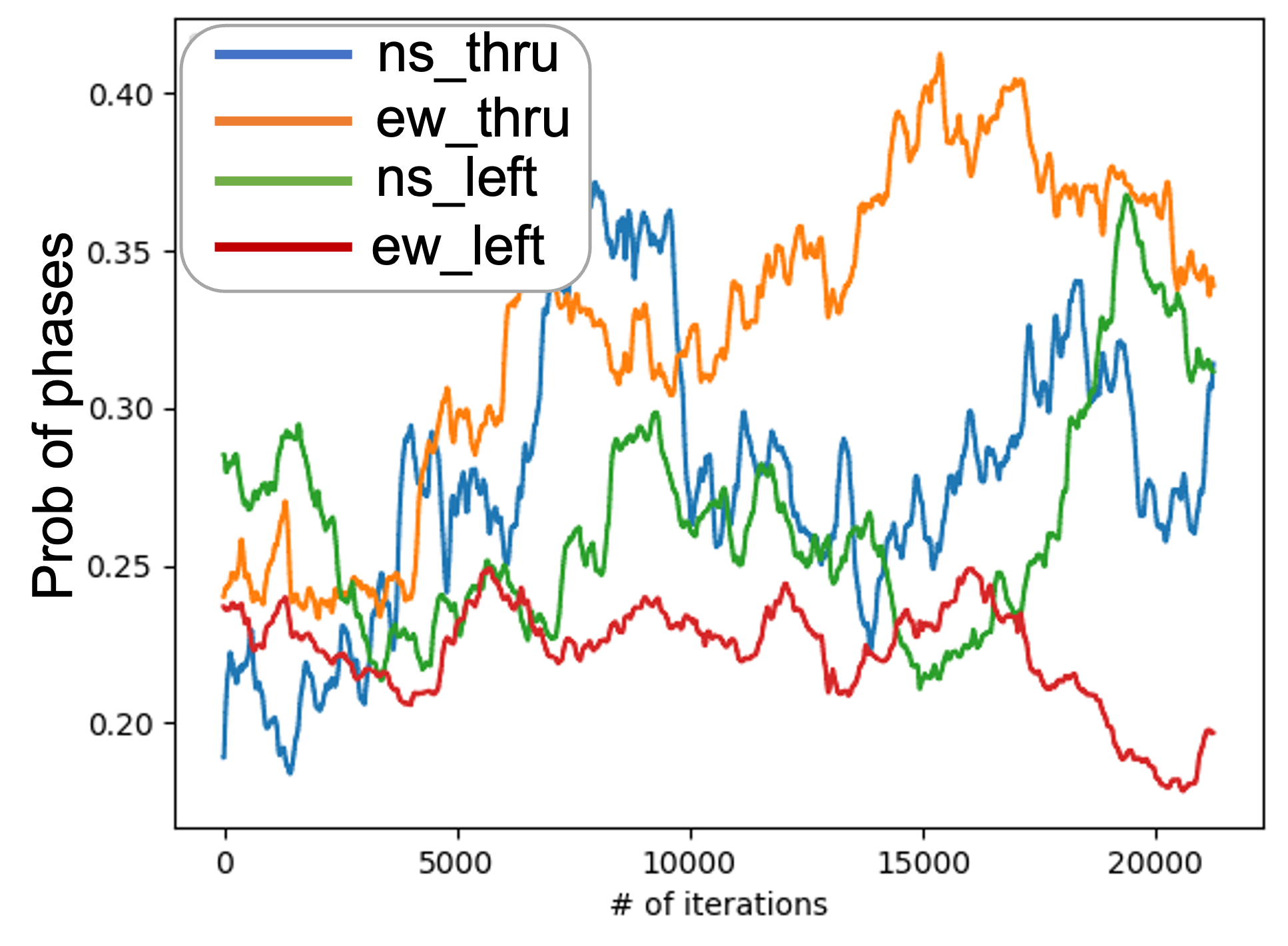}
}
\caption{\blue{Using the PPO algorithm:  ns\_thru, ew\_thru, ns\_left and ew\_left correspond to the four phases for north-south through, east-west through, north-south left-turn, and east-west left turn traffic. The tested intersection traffic state is specifically designed such that a sensible policy must turn the green light for the ew\_left phase. (a) when the training data is in free-flow, a sensible policy is learned. (b) The policy deteriorates when    trained with congested data. }}
\label{ppo-method}
\end{figure}

\section{More realistic routing behaviors}

The DRL learning experiments in this paper assumes full driver adaptation at intersections and vehicles do not have destinations, which is designed to maintain a constant density level and avoid the notorious gridlock issues. In this appendix we provide more evidence that our findings still hold given more realistic routing behaviors and ODs. 

\begin{figure}[!htbp]
\centering
\includegraphics[width=1.0 \linewidth]{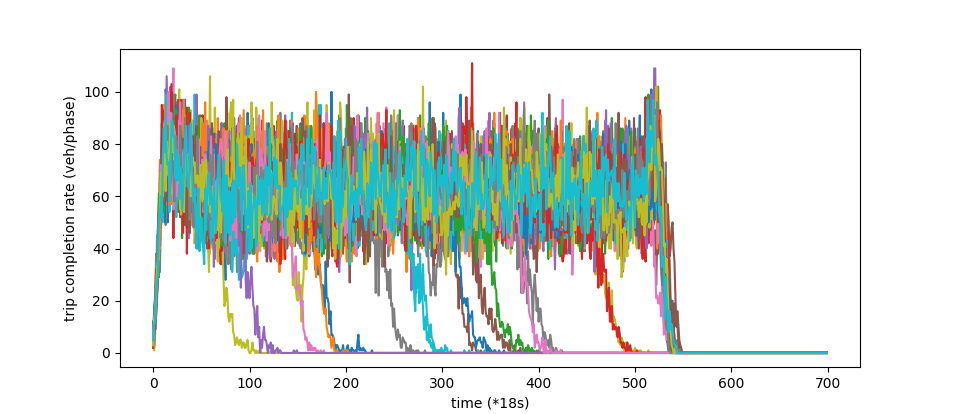}
\caption{Deterioration of the network output under the LQF signal control ( $k=0.2$, $P_r = 0.5$, $T_r = 120s$): each curve corresponds to one simulation.}
\label{gridlock}
\end{figure}

Now all vehicles are equipped with true destinations which are evenly distributed over the network, and human drivers can be adaptive or not according to a probability $P_r$. For adaptive human drivers, it means the route will be recalculated and updated to reduce the travel time based on the dynamic traffic conditions. It needs one more parameter, the rerouting period $T_r$, which indicates how frequently they might update their routes according to dynamic traffic conditions. To see the effect of gridlock on the network output, Fig.\ref{gridlock} depicts the history of completed trips from 50 repetitive simulations under LQF signal control with $k=0.2$, $P_r = 0.5$, $T_r = 120s$. 

\begin{figure}[!htbp]
\centering 
\subfigure[$P_r=0.5$ , $T_r=120s$]{
\includegraphics[width=0.4\linewidth]{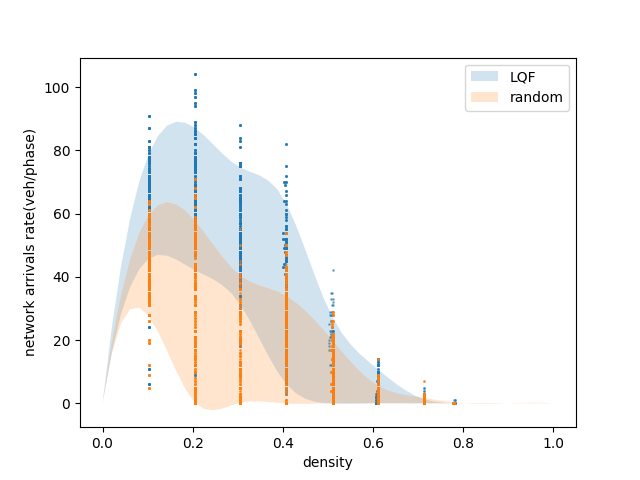}
}
\subfigure[$P_r=1.0$ , $T_r=120s$]{
\includegraphics[width=0.4\linewidth]{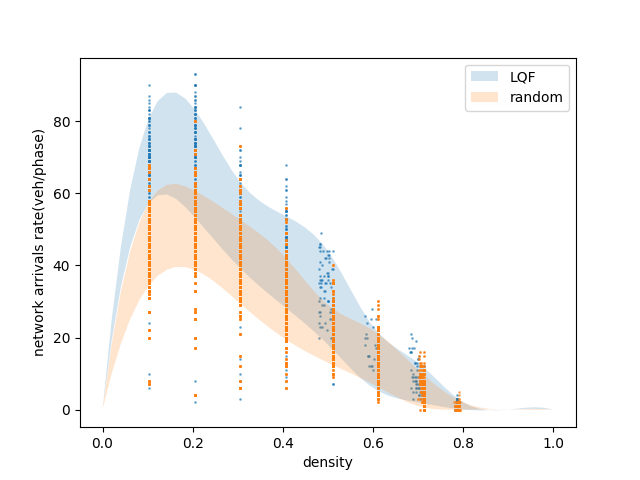}
}
\caption{Comparing LQF and random policies at fixed densities with different driver adaptation: upper and lower bounds correspond to 90\% and 10\% percentiles of the network trip completion rates collected from 200 cycles.}
\label{mfd-adaptation}
\end{figure}

Apparently, realistic routing behaviors can not produce stable network throughput. To derive a sensible MFD,  the network trip completion rates are collected only from first 150 cycles of each simulation to avoid the gridlocks, and the results are summarized in Fig.\ref{mfd-adaptation}. Unlike the throughput we showed in the paper with random turnings, the network throughputs here are all unstable and decay towards zero due to the gridlock.

The simulations with realistic routing behaviors output different MFD shapes under the same signal control policies. However, the new results still support our major finding on the effect of signal control: the LQF outperforms a random policy in light and moderate congestion, but they are almost equivalent in extreme congested scenarios. Thus we conclude the congested network property does not vary with the driver adaptation levels, and research is ongoing to investigate the challenges of non-adaptive drivers to learning methods.




\ifCLASSOPTIONcaptionsoff
  \newpage
\fi



%

\bibliography{tits.bib}

\begin{thebibliography}{10}
\providecommand{\url}[1]{#1}
\csname url@samestyle\endcsname
\providecommand{\newblock}{\relax}
\providecommand{\bibinfo}[2]{#2}
\providecommand{\BIBentrySTDinterwordspacing}{\spaceskip=0pt\relax}
\providecommand{\BIBentryALTinterwordstretchfactor}{4}
\providecommand{\BIBentryALTinterwordspacing}{\spaceskip=\fontdimen2\font plus
\BIBentryALTinterwordstretchfactor\fontdimen3\font minus
  \fontdimen4\font\relax}
\providecommand{\BIBforeignlanguage}[2]{{%
\expandafter\ifx\csname l@#1\endcsname\relax
\typeout{** WARNING: IEEEtran.bst: No hyphenation pattern has been}%
\typeout{** loaded for the language `#1'. Using the pattern for}%
\typeout{** the default language instead.}%
\else
\language=\csname l@#1\endcsname
\fi
#2}}
\providecommand{\BIBdecl}{\relax}
\BIBdecl

\bibitem{daganzo1996nature}
C.~F. Daganzo, ``The nature of freeway gridlock and how to prevent it,''
  \emph{Transportation and traffic theory}, pp. 629--646, 1996.

\bibitem{daganzo1998queue}
------, ``Queue spillovers in transportation networks with a route choice,''
  \emph{Transportation Science}, pp. 3--11, 1998.

\bibitem{nair2001non}
A.~S. Nair, J.-C. Liu, L.~Rilett, and S.~Gupta, ``Non-linear analysis of
  traffic flow,'' in \emph{ITSC 2001. 2001 IEEE Intelligent Transportation
  Systems. Proceedings (Cat. No. 01TH8585)}.\hskip 1em plus 0.5em minus
  0.4em\relax IEEE, 2001, pp. 681--685.

\bibitem{adewumi2016application}
A.~Adewumi, J.~Kagamba, and A.~Alochukwu, ``Application of chaos theory in the
  prediction of motorised traffic flows on urban networks,'' \emph{Mathematical
  Problems in Engineering}, 2016.

\bibitem{nagel1995emergent}
K.~Nagel and M.~Paczuski, ``Emergent traffic jams,'' \emph{Physical Review E},
  vol.~51, no.~4, p. 2909, 1995.

\bibitem{nagatani2002physics}
T.~Nagatani, ``The physics of traffic jams,'' \emph{Reports on progress in
  physics}, vol.~65, no.~9, p. 1331, 2002.

\bibitem{helbing2001traffic}
D.~Helbing, ``Traffic and related self-driven many-particle systems,''
  \emph{Reviews of modern physics}, vol.~73, no.~4, p. 1067, 2001.

\bibitem{chowdhury2000statistical}
D.~Chowdhury, L.~Santen, and A.~Schadschneider, ``Statistical physics of
  vehicular traffic and some related systems,'' \emph{Physics Reports}, vol.
  329, no. 4-6, pp. 199--329, 2000.

\bibitem{nagel2003still}
K.~Nagel, P.~Wagner, and R.~Woesler, ``Still flowing: Approaches to traffic
  flow and traffic jam modeling,'' \emph{Operations research}, vol.~51, no.~5,
  pp. 681--710, 2003.

\bibitem{stauffer2018introduction}
D.~Stauffer and A.~Aharony, \emph{Introduction to percolation theory}.\hskip
  1em plus 0.5em minus 0.4em\relax Taylor \& Francis, 2018.

\bibitem{schroeder2009fractals}
M.~Schroeder, \emph{Fractals, chaos, power laws: Minutes from an infinite
  paradise}.\hskip 1em plus 0.5em minus 0.4em\relax Courier Corporation, 2009.

\bibitem{khamis2014adaptive}
M.~A. Khamis and W.~Gomaa, ``Adaptive multi-objective reinforcement learning
  with hybrid exploration for traffic signal control based on cooperative
  multi-agent framework,'' \emph{Engineering Applications of Artificial
  Intelligence}, vol.~29, pp. 134--151, 2014.

\bibitem{chu2016large}
T.~Chu, S.~Qu, and J.~Wang, ``Large-scale traffic grid signal control with
  regional reinforcement learning,'' in \emph{2016 American Control Conference
  (ACC)}.\hskip 1em plus 0.5em minus 0.4em\relax IEEE, 2016, pp. 815--820.

\bibitem{xu2018network}
M.~Xu, J.~Wu, L.~Huang, R.~Zhou, T.~Wang, and D.~Hu, ``Network-wide traffic
  signal control based on the discovery of critical nodes and deep
  reinforcement learning,'' \emph{Journal of Intelligent Transportation
  Systems}, pp. 1--10, 2018.

\bibitem{ge2019cooperative}
H.~Ge, Y.~Song, C.~Wu, J.~Ren, and G.~Tan, ``Cooperative deep q-learning with
  q-value transfer for multi-intersection signal control,'' \emph{IEEE Access},
  vol.~7, pp. 40\,797--40\,809, 2019.

\bibitem{wei2019colight}
H.~Wei, N.~Xu, H.~Zhang, G.~Zheng, X.~Zang, C.~Chen, W.~Zhang, Y.~Zhu, K.~Xu,
  and Z.~Li, ``Colight: Learning network-level cooperation for traffic signal
  control,'' in \emph{Proceedings of the 28th ACM International Conference on
  Information and Knowledge Management}, 2019, pp. 1913--1922.

\bibitem{tan2019cooperative}
T.~Tan, F.~Bao, Y.~Deng, A.~Jin, Q.~Dai, and J.~Wang, ``Cooperative deep
  reinforcement learning for large-scale traffic grid signal control,''
  \emph{IEEE transactions on cybernetics}, 2019.

\bibitem{gong2019decentralized}
\BIBentryALTinterwordspacing
Y.~Gong, M.~Abdel-Aty, Q.~Cai, and M.~S. Rahman, ``A decentralized network
  level adaptive signal control algorithm by deep reinforcement learning,''
  Tech. Rep., 2019. [Online]. Available:
  \url{http://www.sciencedirect.com/science/article/pii/S259019821930020X}
\BIBentrySTDinterwordspacing

\bibitem{belletti2017expert}
F.~Belletti, D.~Haziza, G.~Gomes, and A.~M. Bayen, ``Expert level control of
  ramp metering based on multi-task deep reinforcement learning,'' \emph{IEEE
  Transactions on Intelligent Transportation Systems}, vol.~19, no.~4, pp.
  1198--1207, 2017.

\bibitem{Herman84}
R.~Herman and S.~Ardekani, ``Characterizing traffic conditions in urban
  areas,'' \emph{Transportation Science}, vol.~18, no.~2, pp. 101--140, 1984.

\bibitem{Hani85}
H.~Mahmassani, R.~Herman, and M.~Walton, ``Characterizing the evolution of
  urban patterns and traffic network performance,'' Tech. Rep., 1985.

\bibitem{mahmassani1990network}
H.~S. Mahmassani, R.~Jayakrishnan, and R.~Herman, ``Network traffic flow
  theory: Microscopic simulation experiments on supercomputers,''
  \emph{Transportation Research Part A: General}, vol.~24, no.~2, pp. 149--162,
  1990.

\bibitem{Hani1985}
J.~Williams, H.~Mahmassani, and R.~Herman, ``\BIBforeignlanguage{English
  (US)}{Analysis of traffic network flow relations and two-fluid model
  parameter sensitivity.}'' \emph{\BIBforeignlanguage{English
  (US)}{Transportation Research Record}}, pp. 95--106, 1 1985.

\bibitem{Laval2015Stochastic}
J.~A. Laval and F.~Castrill{\'o}n, ``Stochastic approximations for the
  macroscopic fundamental diagram of urban networks,'' \emph{Transportation
  Research Procedia}, vol.~7, pp. 615--630, 2015.

\bibitem{gayah2014impacts}
V.~V. Gayah, X.~S. Gao, and A.~S. Nagle, ``On the impacts of locally adaptive
  signal control on urban network stability and the macroscopic fundamental
  diagram,'' \emph{Transportation Research Part B: Methodological}, vol.~70,
  pp. 255--268, 2014.

\bibitem{girault2016exploratory}
J.-T. Girault, V.~V. Gayah, I.~Guler, and M.~Menendez, ``Exploratory analysis
  of signal coordination impacts on macroscopic fundamental diagram,''
  \emph{Transportation Research Record}, vol. 2560, no.~1, pp. 36--46, 2016.

\bibitem{abdelghaffar2019novel}
H.~M. Abdelghaffar and H.~A. Rakha, ``A novel decentralized game-theoretic
  adaptive traffic signal controller: Large-scale testing,'' \emph{Sensors},
  vol.~19, no.~10, p. 2282, 2019.

\bibitem{camponogara2003distributed}
E.~Camponogara and W.~Kraus, ``Distributed learning agents in urban traffic
  control,'' in \emph{Portuguese Conference on Artificial Intelligence}.\hskip
  1em plus 0.5em minus 0.4em\relax Springer, 2003, pp. 324--335.

\bibitem{choi2000hidden}
S.~P. Choi, D.-Y. Yeung, and N.~L. Zhang, ``Hidden-mode markov decision
  processes for nonstationary sequential decision making,'' in \emph{Sequence
  Learning}.\hskip 1em plus 0.5em minus 0.4em\relax Springer, 2000, pp.
  264--287.

\bibitem{da2006dealing}
B.~C. Da~Silva, E.~W. Basso, A.~L. Bazzan, and P.~M. Engel, ``Dealing with
  non-stationary environments using context detection,'' in \emph{Proceedings
  of the 23rd international conference on Machine learning}.\hskip 1em plus
  0.5em minus 0.4em\relax ACM, 2006, pp. 217--224.

\bibitem{daganzo2011macroscopic}
C.~F. Daganzo, V.~V. Gayah, and E.~J. Gonzales, ``Macroscopic relations of
  urban traffic variables: Bifurcations, multivaluedness and instability,''
  \emph{Transportation Research Part B: Methodological}, vol.~45, no.~1, pp.
  278--288, 2011.

\bibitem{jin2013kinematic}
W.-L. Jin, Q.-J. Gan, and V.~V. Gayah, ``A kinematic wave approach to traffic
  statics and dynamics in a double-ring network,'' \emph{Transportation
  Research Part B: Methodological}, vol.~57, pp. 114--131, 2013.

\bibitem{xu2020analytical}
G.~Xu, Z.~Yu, and V.~V. Gayah, ``Analytical method to approximate the impact of
  turning on the macroscopic fundamental diagram,'' \emph{Transportation
  research record}, vol. 2674, no.~9, pp. 933--947, 2020.

\bibitem{SUMO2012}
\BIBentryALTinterwordspacing
D.~Krajzewicz, J.~Erdmann, M.~Behrisch, and L.~Bieker, ``Recent development and
  applications of {SUMO - Simulation of Urban MObility},'' \emph{International
  Journal On Advances in Systems and Measurements}, vol.~5, no. 3\&4, pp.
  128--138, December 2012. [Online]. Available: \url{http://elib.dlr.de/80483/}
\BIBentrySTDinterwordspacing

\bibitem{godfrey1969mechanism}
J.~Godfrey, ``The mechanism of a road network,'' \emph{Traffic Engineering \&
  Control}, vol.~8, no.~8, 1969.

\bibitem{Smeed1967Road}
R.~J. Smeed, ``The road capacity of city centers,'' \emph{Highway Research
  Record}, no. 169, 1967.

\bibitem{Herman1979Two}
R.~Herman and I.~Prigogine, ``A two-fluid approach to town traffic,''
  \emph{Science}, vol. 204, no. 4389, pp. 148--151, 1979.

\bibitem{mahmassani1984investigation}
H.~S. Mahmassani, J.~C. Williams, and R.~Herman, ``Investigation of
  network-level traffic flow relationships: some simulation results,''
  \emph{Transportation Research Record}, vol. 971, pp. 121--130, 1984.

\bibitem{Lighthill1955Kinematic}
M.~J. Lighthill and G.~B. Whitham, ``On kinematic waves ii. a theory of traffic
  flow on long crowded roads,'' \emph{Proceedings of the Royal Society of
  London. Series A. Mathematical and Physical Sciences}, vol. 229, no. 1178,
  pp. 317--345, 1955.

\bibitem{richards1956shock}
P.~I. Richards, ``Shock waves on the highway,'' \emph{Operations research},
  vol.~4, no.~1, pp. 42--51, 1956.

\bibitem{daganzo2008analytical}
C.~F. Daganzo and N.~Geroliminis, ``An analytical approximation for the
  macroscopic fundamental diagram of urban traffic,'' \emph{Transportation
  Research Part B: Methodological}, vol.~42, no.~9, pp. 771--781, 2008.

\bibitem{mnih2015human}
V.~Mnih, K.~Kavukcuoglu, D.~Silver, A.~A. Rusu, J.~Veness, M.~G. Bellemare,
  A.~Graves, M.~Riedmiller, A.~K. Fidjeland, G.~Ostrovski \emph{et~al.},
  ``Human-level control through deep reinforcement learning,'' \emph{Nature},
  vol. 518, no. 7540, p. 529, 2015.

\bibitem{silver2017mastering}
D.~Silver, J.~Schrittwieser, K.~Simonyan, I.~Antonoglou, A.~Huang, A.~Guez,
  T.~Hubert, L.~Baker, M.~Lai, A.~Bolton \emph{et~al.}, ``Mastering the game of
  go without human knowledge,'' \emph{Nature}, vol. 550, no. 7676, p. 354,
  2017.

\bibitem{chen2019model}
J.~Chen, B.~Yuan, and M.~Tomizuka, ``Model-free deep reinforcement learning for
  urban autonomous driving,'' in \emph{2019 IEEE intelligent transportation
  systems conference (ITSC)}.\hskip 1em plus 0.5em minus 0.4em\relax IEEE,
  2019, pp. 2765--2771.

\bibitem{li2016traffic}
L.~Li, Y.~Lv, and F.-Y. Wang, ``Traffic signal timing via deep reinforcement
  learning,'' \emph{IEEE/CAA Journal of Automatica Sinica}, vol.~3, no.~3, pp.
  247--254, 2016.

\bibitem{genders2016using}
W.~Genders and S.~Razavi, ``Using a deep reinforcement learning agent for
  traffic signal control,'' \emph{arXiv preprint arXiv:1611.01142}, 2016.

\bibitem{chu2015traffic}
T.~Chu and J.~Wang, ``Traffic signal control with macroscopic fundamental
  diagrams,'' in \emph{2015 American Control Conference (ACC)}.\hskip 1em plus
  0.5em minus 0.4em\relax IEEE, 2015, pp. 4380--4385.

\bibitem{chu2019multi}
T.~Chu, J.~Wang, L.~Codec{\`a}, and Z.~Li, ``Multi-agent deep reinforcement
  learning for large-scale traffic signal control,'' \emph{IEEE Transactions on
  Intelligent Transportation Systems}, 2019.

\bibitem{willianms1988toward}
R.~Willianms, ``Toward a theory of reinforcement-learning connectionist
  systems,'' \emph{Technical Report NU-CCS-88-3, Northeastern University},
  1988.

\bibitem{bellman1957markovian}
R.~Bellman, ``A markovian decision process,'' \emph{Journal of mathematics and
  mechanics}, pp. 679--684, 1957.

\bibitem{bertsekas1987dynamic}
D.~P. Bertsekas, \emph{Dynamic Programming: Determinist. and Stochast.
  Models}.\hskip 1em plus 0.5em minus 0.4em\relax Prentice-Hall, 1987.

\bibitem{howard1960dynamic}
R.~A. Howard, ``Dynamic programming and markov processes.'' 1960.

\bibitem{puterman1994markovian}
M.~Puterman, ``Markovian decision problems,'' 1994.

\bibitem{sutton1999policy}
R.~Sutton, \emph{The policy process: an overview}.\hskip 1em plus 0.5em minus
  0.4em\relax Overseas Development Institute London, 1999.

\bibitem{lillicrap2015continuous}
T.~P. Lillicrap, J.~J. Hunt, A.~Pritzel, N.~Heess, T.~Erez, Y.~Tassa,
  D.~Silver, and D.~Wierstra, ``Continuous control with deep reinforcement
  learning,'' \emph{arXiv preprint arXiv:1509.02971}, 2015.

\bibitem{kheterpal2018flow}
N.~Kheterpal, K.~Parvate, C.~Wu, A.~Kreidieh, E.~Vinitsky, and A.~Bayen,
  ``Flow: Deep reinforcement learning for control in sumo,'' \emph{SUMO}, pp.
  134--151, 2018.

\bibitem{vinitsky2018benchmarks}
E.~Vinitsky, A.~Kreidieh, L.~Le~Flem, N.~Kheterpal, K.~Jang, F.~Wu, R.~Liaw,
  E.~Liang, and A.~M. Bayen, ``Benchmarks for reinforcement learning in
  mixed-autonomy traffic,'' in \emph{Conference on Robot Learning}, 2018, pp.
  399--409.

\bibitem{arel2010reinforcement}
I.~Arel, C.~Liu, T.~Urbanik, and A.~Kohls, ``Reinforcement learning-based
  multi-agent system for network traffic signal control,'' \emph{IET
  Intelligent Transport Systems}, vol.~4, no.~2, pp. 128--135, 2010.

\bibitem{dai2011neural}
Y.~Dai, J.~Hu, D.~Zhao, and F.~Zhu, ``Neural network based online traffic
  signal controller design with reinforcement training,'' in \emph{2011 14th
  International IEEE Conference on Intelligent Transportation Systems
  (ITSC)}.\hskip 1em plus 0.5em minus 0.4em\relax IEEE, 2011, pp. 1045--1050.

\bibitem{geroliminis2012optimal}
N.~Geroliminis, J.~Haddad, and M.~Ramezani, ``Optimal perimeter control for two
  urban regions with macroscopic fundamental diagrams: A model predictive
  approach,'' \emph{IEEE Transactions on Intelligent Transportation Systems},
  vol.~14, no.~1, pp. 348--359, 2012.

\bibitem{haddad2012stability}
J.~Haddad and N.~Geroliminis, ``On the stability of traffic perimeter control
  in two-region urban cities,'' \emph{Transportation Research Part B:
  Methodological}, vol.~46, no.~9, pp. 1159--1176, 2012.

\bibitem{aboudolas2013perimeter}
K.~Aboudolas and N.~Geroliminis, ``Perimeter and boundary flow control in
  multi-reservoir heterogeneous networks,'' \emph{Transportation Research Part
  B: Methodological}, vol.~55, pp. 265--281, 2013.

\bibitem{newell2002simplified}
G.~F. Newell, ``A simplified car-following theory: a lower order model,''
  \emph{Transportation Research Part B: Methodological}, vol.~36, no.~3, pp.
  195--205, 2002.

\bibitem{Dag04a}
C.~F. Daganzo, ``In traffic flow, cellular automata = kinematic waves,''
  \emph{Transportation Research Part B}, vol.~40, no.~5, pp. 396--403, 2006.

\bibitem{Lav16}
J.~A. Laval and B.~R. Chilukuri, ``Symmetries in the kinematic wave model and a
  parameter-free representation of traffic flow,'' \emph{Transportation
  Research Part B: Methodological}, vol.~89, pp. 168--177, 2016.

\bibitem{wolfram1984cellular}
S.~Wolfram, ``Cellular automata as models of complexity,'' \emph{Nature}, vol.
  311, no. 5985, p. 419, 1984.

\bibitem{kuurkova1992kolmogorov}
V.~Kurkov{\'a}, ``Kolmogorov's theorem and multilayer neural networks,''
  \emph{Neural networks}, vol.~5, no.~3, pp. 501--506, 1992.

\bibitem{zhou2021gridlock}
H.~Zhou and J.~A. Laval, ``Avoiding gridlock on large congested networks: A
  multi-agent deep reinforcement learning approach with spillback knowledge,''
  \emph{Transportation Research Board 100th Annual MeetingTransportation
  Research Board}, no. TRBAM-21-04245, 2021.

\bibitem{sutton2018reinforcement}
R.~S. Sutton and A.~G. Barto, \emph{Reinforcement learning: An
  introduction}.\hskip 1em plus 0.5em minus 0.4em\relax MIT press, 2018.

\bibitem{sutton1988learning}
R.~S. Sutton, ``Learning to predict by the methods of temporal differences,''
  \emph{Machine learning}, vol.~3, no.~1, pp. 9--44, 1988.

\bibitem{schulman2017proximal}
J.~Schulman, F.~Wolski, P.~Dhariwal, A.~Radford, and O.~Klimov, ``Proximal
  policy optimization algorithms,'' \emph{arXiv preprint arXiv:1707.06347},
  2017.

\end{thebibliography}
\bibliographystyle{IEEEtran}






\end{document}